\documentclass[10pt,twocolumn,letterpaper]{article}

\usepackage{cvpr}
\usepackage{times}
\usepackage{epsfig}
\usepackage{graphicx}
\usepackage{amsmath}
\usepackage{amssymb}

\newcommand{\real}{\ensuremath{\mathbb{R}}}

\newtheorem{lemma}{Lemma}

\DeclareMathOperator{\Tr}{Tr}
\DeclareMathOperator*{\argmax}{arg\,max}
\DeclareMathOperator*{\argmin}{arg\,min}
\DeclareMathOperator{\vect}{vec}

\usepackage{subfig}
\usepackage{algorithm}
\usepackage{algorithmicx}
\usepackage{algpseudocode}
\usepackage[pagebackref=true,breaklinks=true,letterpaper=true,colorlinks,bookmarks=false]{hyperref}

\cvprfinalcopy 

\def\cvprPaperID{3} 

\ifcvprfinal\fi
\begin{document}

\title{Representations, Metrics and Statistics for Shape Analysis of Elastic Graphs}

\author{Xiaoyang Guo, Anuj Srivastava\\
Department of Statistics, Florida State University\\
Tallahassee, FL 32306, USA\\
{\tt\small \{xiaoyang.guo,anuj\}@stat.fsu.edu}
}

\maketitle

\begin{abstract}
Past approaches for statistical shape analysis of objects have focused mainly on objects within the same topological classes, \eg , scalar functions, 
Euclidean curves, or surfaces, \etc. 
For objects that differ in more complex ways, the current literature offers only topological 
methods. This paper introduces a far-reaching
geometric approach for analyzing shapes of graphical objects,
such as road networks, blood vessels, brain fiber tracts, \etc. It represents such objects, exhibiting differences in both
geometries and topologies, as graphs made of
curves with arbitrary shapes (edges) and connected at arbitrary junctions (nodes). 

To perform statistical analyses, one needs mathematical representations,  metrics and other geometrical tools, such as geodesics, means, and covariances. 
This paper utilizes a quotient structure to develop efficient algorithms 
for computing these quantities, leading to useful statistical tools, including principal component analysis and 
analytical statistical testing and modeling of graphical shapes. The efficacy of this framework is demonstrated using various simulated 
as well as the real data from neurons and brain arterial networks.
\end{abstract}

\section{Introduction}

The problem of analyzing shapes of objects has steadily gained in importance over the last few years~\cite{dryden2016statistical,jermyn2017elastic, laga2018survey, srivastava2016functional}. 
This rise is fueled by the availability of multimodal, high-dimensional data that records objects of interest in a 
variety of contexts and applications. Shapes of objects help characterize their identity, classes, movements, and 
roles in larger scenes. Consequently, many approaches have been developed for 
comparing, summarizing, modeling, testing, and tracking shapes in static image or video data. While 
early methods generally relied on discrete representations of objects (point clouds, landmarks)~\cite{kendall1984shape}, more
recent methods have focused on continuous objects such as functions~\cite{srivastava2011registration}, curves~\cite{klassen2004analysis}, and surfaces~\cite{kurtek2010novel}. The main 
motivation for this paradigm shift comes from the need to address {\it registration}, considered the most challenging issue in shape analysis. 
Registration refers to establishing a correspondence between points or features across objects and is an important
ingredient in comparing shapes.
Continuous representations of objects use convenient actions of the parameterization groups to help 
solve dense registration problems~\cite{srivastava2016functional}. Furthermore, they use elastic Riemannian metrics -- which are 
invariant to the actions of re-parameterization groups -- and some simplifying square-root representations, to develop 
very efficient techniques for comparing and analyzing shapes. 

\begin{figure}
\begin{center}
\subfloat[Brain Artery~\cite{bullitt2005vessel}]{
\includegraphics[width=0.52\linewidth]{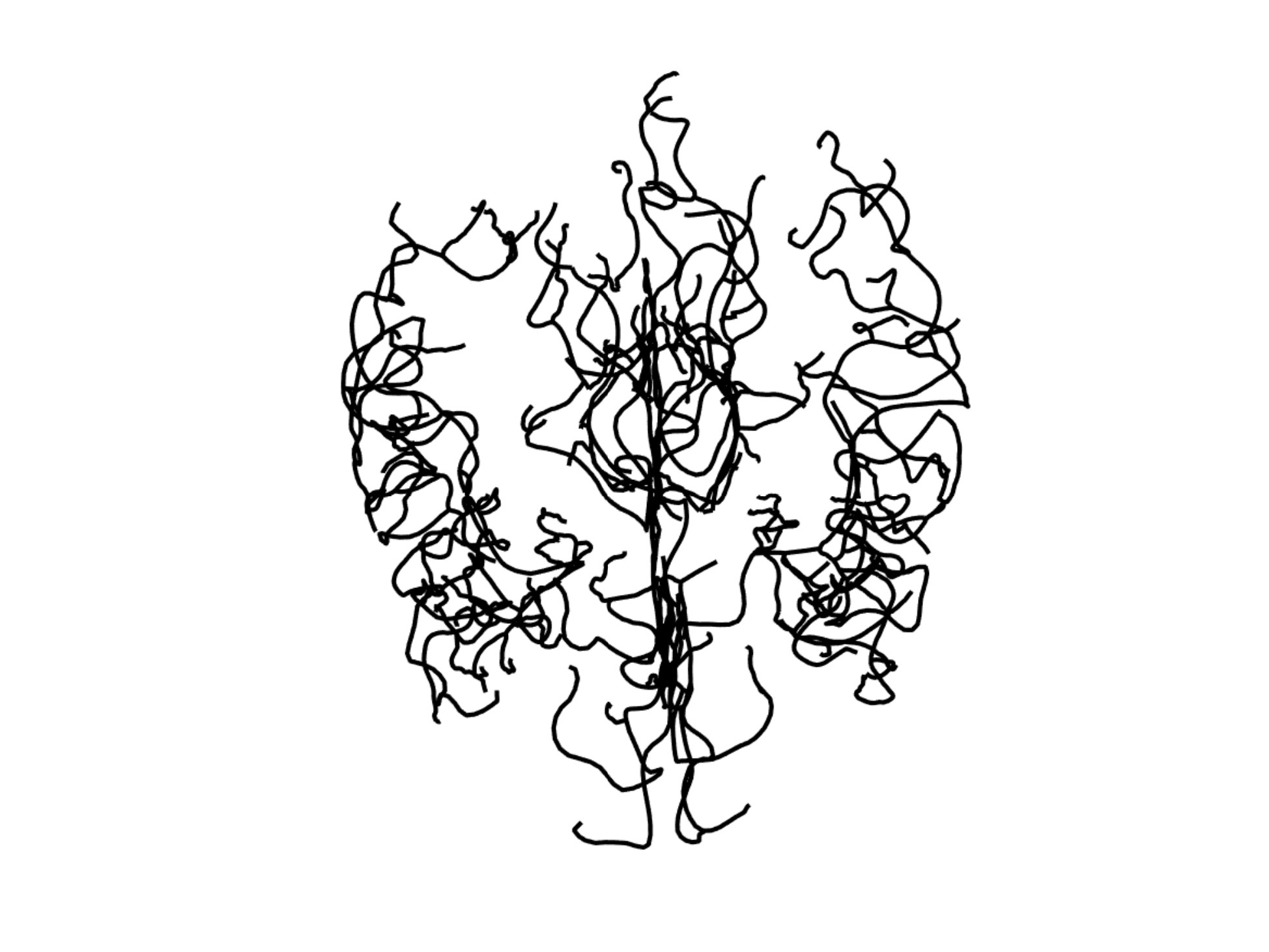}}
\subfloat[Retinal Blood Vessel~\cite{hoover2000locating}]{
\includegraphics[width=0.45\linewidth]{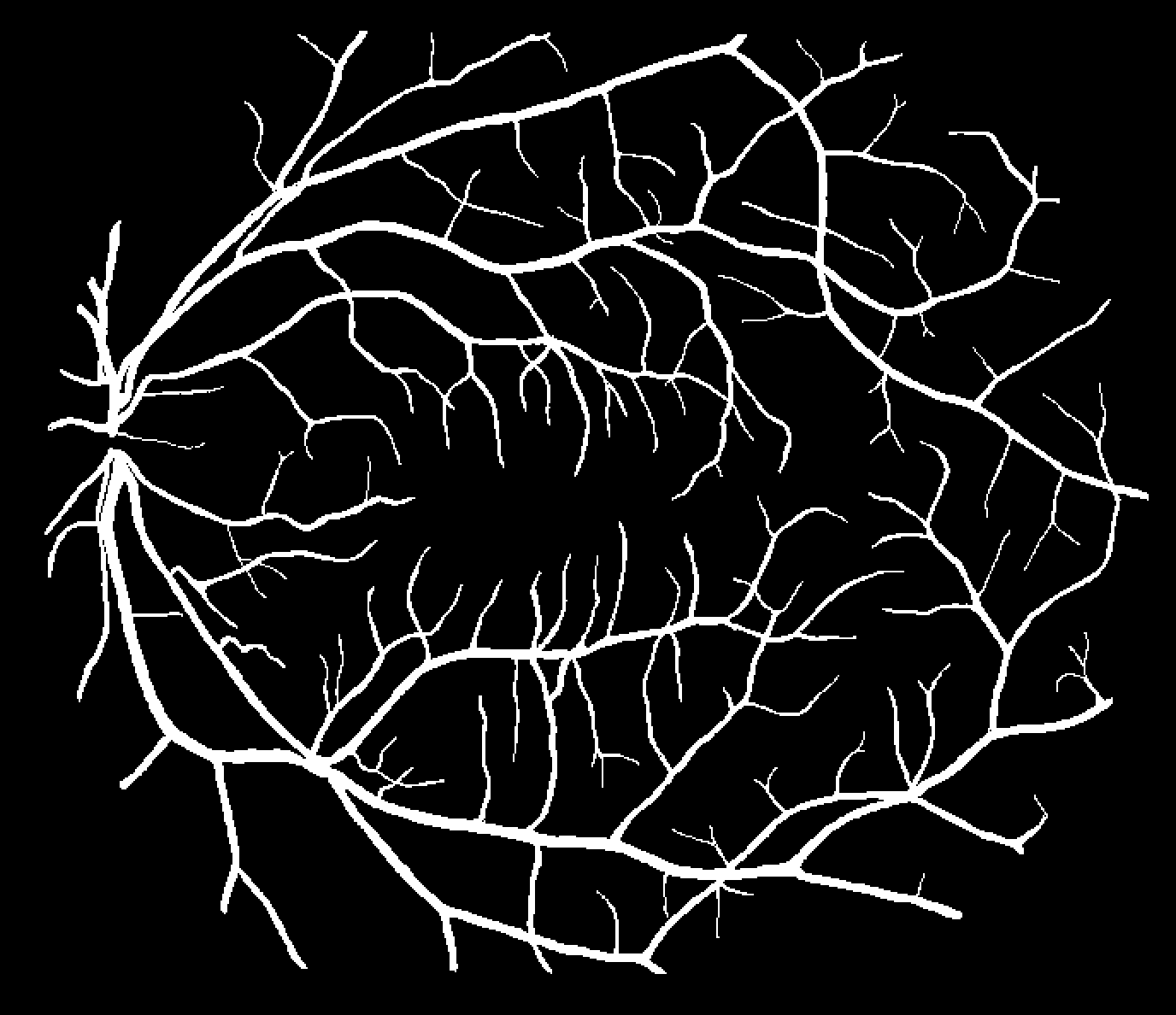}}\\
\subfloat[Fruit Fly Wing~\cite{sonnenschein2015image}]{
\includegraphics[width=0.52\linewidth]{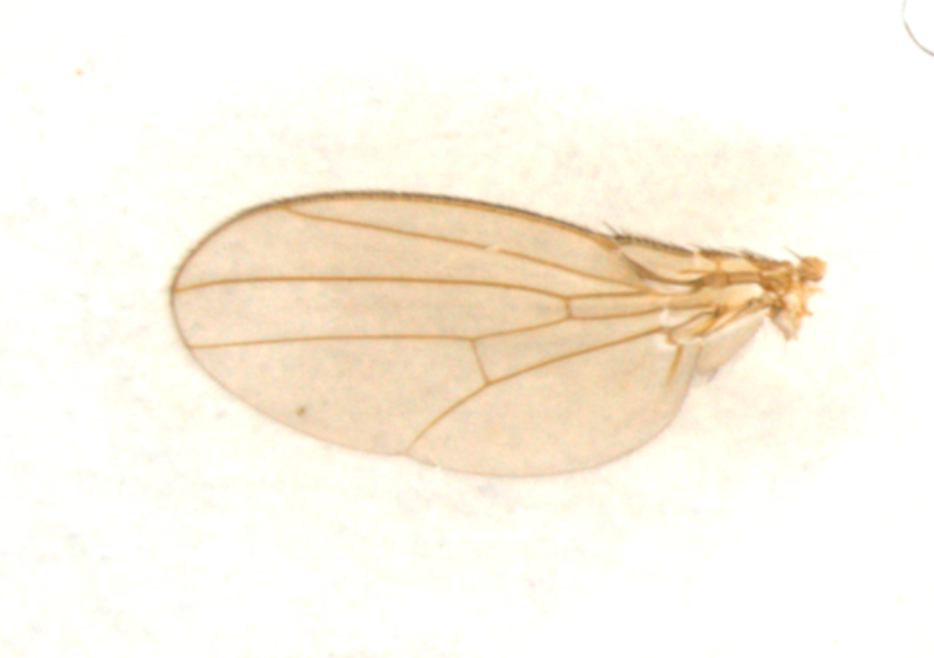}}
\subfloat[Neuron~\cite{kong2005diversity}]{
\includegraphics[width=0.45\linewidth]{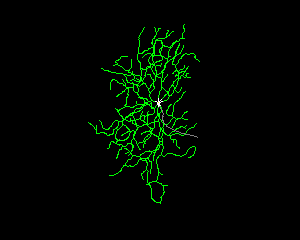}}
\end{center}
\caption{Graphical shapes:  complex networks formed by branching, intersections or 
merging of curves (edges) at arbitrary points (nodes).}
\label{fig:example}
\end{figure}

While elastic shape analysis is considered well developed for some simpler objects -- Euclidean 
curves~\cite{srivastava2010shape}, manifold-valued curves~\cite{zhang2018phase}, 3D surfaces~\cite{jermyn2012elastic, su2019shape}, and simple trees~\cite{duncan2018statistical} --- the problem of analyzing more complex
objects remains elusive. Stated differently, the past developments have mainly focused on objects that 
exhibit only the geometrical 
variabilities in shapes, while being of same or similar topologies. Similar topologies help 
pose the registration problem as that of optimal diffeomorphic re-parameterization of the common domain
(of parameterizations). In this paper we are concerned with comparing objects with potentially very different 
topologies and geometries. Examples of such objects include road networks, vein structures in leaves, network of blood
vessels in human brain or eyes, complex biomolecules with arbitrary branchings and foldings and so on.
Some illustrations of such objects are shown in Fig.  \ref{fig:example}.
A common characteristic of these objects is that they are made of a number of curves, with arbitrary shapes
and sizes, that merge 
and branch at arbitrary junctions, resulting in complex patterns of pathways. In order to compare any two 
such objects one needs to take into account the numbers, locations, and shapes of individual curves. In particular, 
one has to solve a difficult problem of registration of points and features across curves and full objects. \\

\noindent {\bf Topological Data Analysis (TDA) and Its Limitations}: 
How have such shapes been studied in the past? A common and relatively easy approach is to extract 
certain features of interest from each shape, and compare these features across objects using 
some appropriate metrics. One example of this idea is {\it topological data analysis}, where one 
extracts certain mathematical features (\eg, {\it Betti curves}) from the objects and compares these
features using chosen metrics~\cite{bubenik2015statistical,hang2019topological,singh2014topological,wasserman2018topological}. However, the difficulty in such approaches is that these feature-based representations
are typically not invertible. Feature extraction represents many-to-one mapping
(from the object space to a feature space) and it is not clear as to which shapes share 
the same topological representation. Because of the lack of invertibility of representation, it is difficult to
 map statistical quantities back to the object space. 
 \\
 
\noindent{\bf Specific Goals}: In this paper, our goals are to develop tools for a comprehensive statistical analysis of complex
shapes with graphical/network structures. Specifically, we seek: (1) a {\bf shape metric} that is invariant to the
usual shape-preserving transformations, (2)  {\bf registration} of points across objects, (3) computation
of {\bf geodesic} paths between given shapes and (4) computation of {\bf statistical summaries} -- mean, covariance, PCA, 
\etc -- in the shape space of such objects. These tools, in turn, can be used for analysis, clustering, classification,  
and modeling of shapes in conjunction with other machine learning methods. We reiterate that current TDA techniques 
can not provide several of these solutions. 

Our approach is to 
view the objects of interest as graphs (we also call them \textit{elastic graphs}) -- with edges defined by shapes of curves and 
nodes by the junctions of these curves. These graphs are then
represented by their adjacency matrices, with elements given by the shapes of the corresponding 
edges. Since the ordering of nodes in these graphs and indices in the associated adjacency matrices, is 
arbitrary, we model this variability using an action of the permutation group, and 
represent each graph as an orbit under this group. Then, we develop techniques for optimization under this 
permutation group (also known as {\it graph matching}), computing geodesics and summaries under the induced
metric on the Riemannian quotient space, termed as the {\it graph shape space}. The main contributions of 
this paper as follows. There is no currently existing geometrical framework for shape analysis of such graphical 
objects. While TDA and other such methods can provide a measure of dissimilarity in shapes, this paper 
provides statistical quantities such as mean, covariance, principal modes, \etc, for a more comprehensive shape
summarization and modeling. 

The rest of the paper is organized as follows. 
Section \ref{sec:framework} introduces the mathematical framework for analyzing graphical shapes.
In Section \ref{sec:graphmatching} we discuss the graph matching problem,
 followed by computations of shape summaries in Section \ref{sec:shapesummary}.
Sections \ref{sec: neuron} and \ref{sec: brain artery tree} present applications on neurons and brain arteries.
In the end, Section \ref{sec:conclusion} concludes the paper.

\section{Proposed Mathematical Framework}
\label{sec:framework}

We now present a mathematical framework for representing graph objects. 
The proposed framework can be viewed as an extension of some previous works on graphs
~\cite{calissano2020populations,guo2019quotient,jain2009structure,jain2011graph,jain2012learning}. 
However, those past works restricted to only scalar-valued weighted graphs 
while we are now considering full shapes. 

\subsection{Elastic Graph Representation}
We are interested in studying 
objects that are made of a number of curves, with arbitrary shapes and placements, that merge 
and branch at arbitrary junctions, resulting in complex networks of pathways. 
We will represent them as graphs with nodes corresponding to junctions and edges 
corresponding to the shapes of curves connecting the nodes. Here we assume that any two nodes
are connected by at most one curve. 
An edge attributed graph $G$ is an ordered pair $(V,a)$, 
where $V$ is a set of nodes and $a$ is an edge attribute function: $a: V \times V \rightarrow {\cal S}$.
(${\cal S}$ is the shape space of elastic Euclidean curves \cite{srivastava2010shape} and is briefly summarized in the appendix.)
The shape $a(v_i, v_j)$ characterizes the curve between nodes $v_i, v_j \in V, i \neq j$.
Assuming that the number of nodes, denoted by $|V|$, 
is  $n$, $G$ can be represented by its adjacency matrix 
$A = \{a_{ij}\} \in {\cal S}^{n \times n}$, where the element $a_{ij} = a(v_i, v_j)$. 
For an undirected graph $G$, we have $a(v_i, v_j) = a(v_j, v_i)$ and therefore $A$ is a symmetric matrix. 
The set of all such matrices is given by ${\cal A} = \{ A \in {\cal S}^{n \times n} | A = A^T, \text{diag}(A) = \mathbf{0}\}$.
(Here $\mathbf{0}$ denotes a null edge.) 
Let $d_s$ denote the shape (Riemannian) distance on ${\cal S}$. We will use this to impose a metric on the 
representation space ${\cal A}$. That is, for any two $A_1, A_2 \in {\cal A}$, 
with the corresponding entries $a_{ij}^1$ and $a_{ij}^2$, respectively, the 
metric: 
\begin{equation}
d_a(A_1, A_2) \equiv
\sqrt{\sum_{i,j} d_s(a_{ij}^1, a_{ij}^2)^2} \ , 
\label{eq:preshape-metric}
\end{equation} 
quantifies the differences between graphs $A_1$ and $A_2$. 
Under the chosen metric, the geodesic or the shortest path between two  points in ${\cal A}$
can be written as a set of geodesics in ${\cal S}$ between the 
corresponding components. That is, for any $A_1, A_2 \in {\cal A}$, the geodesic $\alpha: [0,1] \to {\cal A}$ 
consists of components $\alpha = \{\alpha_{ij} \}$ 
given by $\alpha_{ij}: [0,1] \to {\cal S}$, a uniform-speed geodesic path in $\mathcal{S}$ between $a_{ij}^1$ and $a_{ij}^2$.

The ordering of nodes in graphs is arbitrary, and this variability complicates the ensuing analysis. 
As the result, we need to remove the ordering variability, \ie , the nodes of graphs should be registered via permutation.
A permutation matrix is a matrix that has exactly one 1 in each row and column, with all the other entries being zero. 
Let $\mathcal{P}$ be the group of all $n \times n$ permutation matrices with group operation being matrix multiplication and identity element being the $n \times n$ identity matrix.
We define the action of $\mathcal{P}$ on $\mathcal{A}$ as:
\begin{equation}
\mathcal{P} \times \mathcal{A} \rightarrow \mathcal{A}, P*A = P \cdot A \cdot P^T \ .
\label{eqn:action}
\end{equation}
Here $\cdot$ implies a permutation of entries of $A$ according to the nonzero 
elements of $P$. The full action $P*A$ results in the swapping of rows and columns 
of $A$ according to $P$. It can be shown that this mapping is a proper group action of 
$P$ on ${\cal A}$. 
The orbit of an $A \in {\cal A}$ under the action of $\mathcal{P}$ is given by:
$[A] = \{ PAP^T |P \in \mathcal{P} \}$.
Any two elements of an orbit denote exactly the same graph shape, except that the ordering of the nodes has been changed.
The membership of an orbit defines an equivalent relationship $\sim$ on  ${\cal A}$:
$$
A_1 \sim A_2 \Leftrightarrow \exists P \in \mathcal{P}: P \cdot A_1 \cdot P^T = A_2 \ .
$$
The set of all equivalence classes forms the quotient space or the {\it graph shape space}:
$\mathcal{G} \equiv\mathcal{A} \slash \mathcal{P} = \{[A]|A\in \mathcal{A}\}$.

\begin{lemma}
\begin{enumerate}
\item The action of ${\cal P}$ on the set ${\cal A}$ given in Eqn. \ref{eqn:action} is by isometries. 
That is, for any $P \in {\cal P}$ and $A_1, A_2 \in {\cal A}$, we have 
$d_a(A_1, A_2) = d_a(P*A_1, P*A_2)$.
\item Since this action is isometric and the group ${\cal P}$ is finite, we define a metric on the quotient 
space ${\cal G}$: 
\begin{equation}
\begin{split}
d_g([A_1], [A_2]) & = \min_{P \in {\cal P}} d_a(A_1, P*A_2)  \\
& = \min_{P \in {\cal P}} d_a(A_2, P*A_1)
\label{eq:metric}
\end{split} \ .
\end{equation}
\end{enumerate}
\end{lemma}

Let $\hat{P} = \operatorname*{argmin}_{P \in {\cal P}} d_a(A_1, P*A_2)$, then $A_1$ and $\hat{P}*A_2$ 
are considered to be {\it registered}. A geodesic between $[A_1]$ and $[A_2]$ under 
the metric $d_g$ is given by $[\alpha(t)]$ where
$\alpha: [0,1] \to {\cal A}$ is a geodesic between $A_1$ and $\hat{P}*A_2$. 

\subsection{Graph Matching}
\label{sec:graphmatching}

The problem of optimization over $\mathcal{P}$, stated in Eqn. \ref{eq:metric}, 
is known as the {\it graph matching problem} in the literature. In the simpler special case
where $\mathcal{A}$ is a Euclidean space and edge similarity is measured by 
the Euclidean norm,  the problem can be formulated as $\hat{P} = \argmin_{P\in\mathcal{P}}\|A_1-PA_2P^T\|=\argmax_{P\in\mathcal{P}}\Tr(A_1PA_2P^T)$. 
This particular formulation is called
Koopmans-Beckmann's quadratic assignment programming (QAP) problem~\cite{koopmans1957assignment}.
One can use several existing 
solutions for approximating the optimal registration~\cite{caelli2004eigenspace,liu2012extended,umeyama1988eigendecomposition,vogelstein2015fast}.

When $\mathcal{A}$ represents a more general space, \eg , shape space $\mathcal{S}$ in this paper, 
some of the previous solutions are not applicable.
Instead, the problem can be rephrased as 
$\hat{P} = \argmax_{P \in \mathcal{P}} \vect{(P)}^TK\vect{(P)}$,
where $K \in \real^{n^2 \times n^2}$ is called an {\it affinity matrix}, 
whose entries $k_{aibj}$ measures the affinity between edge $ab$ of $A_1$ and edge $ij$ of $A_2$. 
In this paper we use the shape similarity between two edges, obtained using the square-root velocity function (SRVF) representations 
\cite{srivastava2010shape} while modding out the rotation and the re-parametrization groups, as affinity: 
$$
k_{aibj}=\begin{cases}
0,\  \text{if $a^1_{ab}$ or $a^2_{ij}$ is null} \\
\sup_{O,\gamma}\langle q_1, O(q_2 \circ \gamma)\sqrt{\dot{\gamma}} \rangle,\ \text{otherwise}
\end{cases} \ .
$$
Here $q_1, q_2$ denote the SRVFs of the edges $ab$ of $A_1$ and $ij$ of $A_2$.
$O$ is a rotation matrix and $\gamma$ is a diffeomorphic reparameterization. (Please refer to appendix for details.)
This formulation is called the Lawler's QAP probelm~\cite{lawler1963quadratic}.
For this formulation, there are several algorithms available for approximating the solution
~\cite{cour2007balanced,gold1996graduated,leordeanu2005spectral,leordeanu2009integer,zanfir2018deep,zhou2012factorized,zhou2015factorized}.
In this paper, we use the factorized graph matching (FGM) algorithm~\cite{zhou2015factorized} to match elastic graphs.

So far we have assumed that the graphs being matched are all of the same size (in terms of the 
number of nodes). 
For graphs $G_1$ and $G_2$, with different number of nodes $n_1$ and $n_2$, we 
can pad them using $n_2$ and $n_1$ null nodes, respectively, to bring them to the same size $n_1 + n_2$. 
By doing so, the original (real) nodes of both $G_1$ and $G_2$ can potentially be registered to null nodes in the other graph.

\begin{figure}
\begin{center}
\subfloat[Geodesic in $\mathcal{A}$]{
\includegraphics[width=\linewidth]{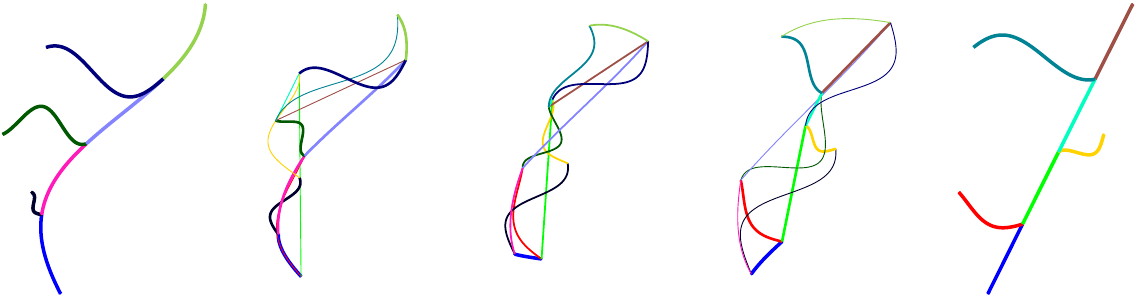}} \\
\subfloat[Geodesic in $\mathcal{G}$]{
\includegraphics[width=\linewidth]{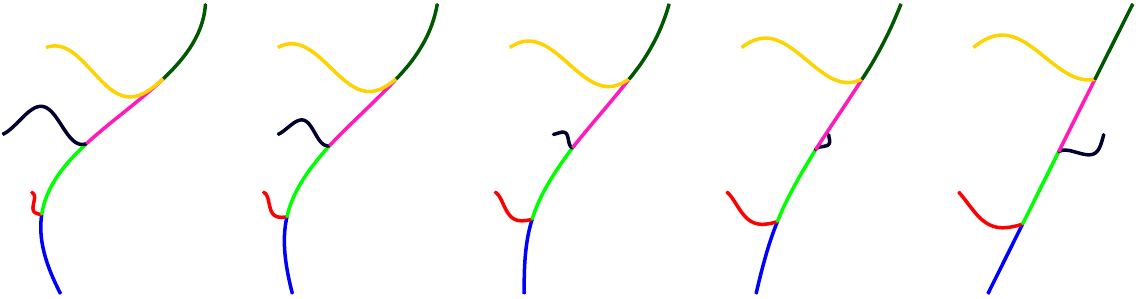}}
\end{center}
\caption{Example of geodesics in adjacency matrix space $\mathcal{A}$ (top) and graph shape space $\mathcal{G}$ (bottom). 
Colors denote registered edges across graphs.}
\label{fig:geo_simu_1}
\end{figure}

\begin{figure}
\begin{center}
\subfloat[Geodesic in $\mathcal{A}$]{
\includegraphics[width=\linewidth]{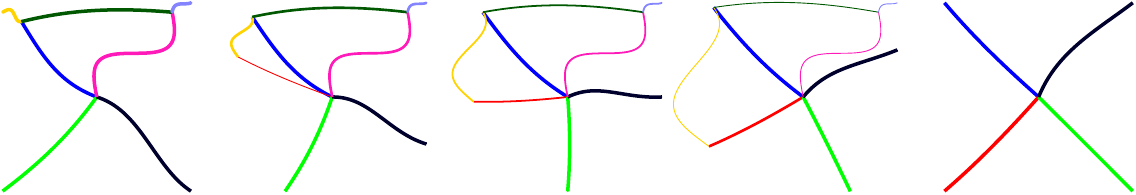}}\\
\subfloat[Geodesic in $\mathcal{G}$]{
\includegraphics[width=\linewidth]{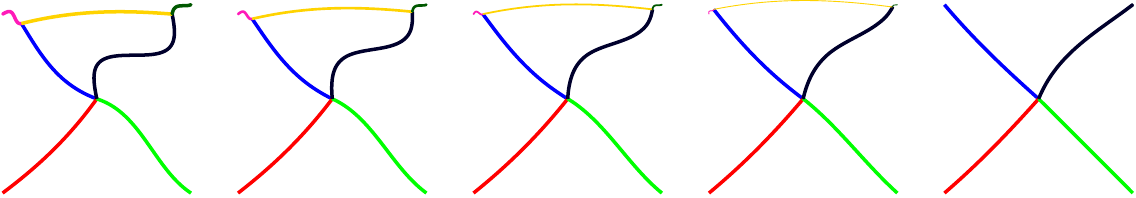}}
\end{center}
\caption{Same as Fig. \ref{fig:geo_simu_1}.}
\label{fig:geo_simu_2}
\end{figure}

We present a couple of simple illustrative examples of this framework in 
Fig. \ref{fig:geo_simu_1} and \ref{fig:geo_simu_2}. 
In each case we show two graphs $G_1$ and $G_2$ drawn as the first and the last graphs in each picture. 
Then, we show a geodesic path between them in two different spaces -- ${\cal A}$ and ${\cal G}$, \ie , without and 
with registration. The deformations between registered graphs, associated with 
geodesics in ${\cal G}$,  look much more natural than those in ${\cal A}$.  The edge features are
preserved better in the intermediate graphs along the geodesics in ${\cal G}$. 

\section{Shape Summaries of Elastic Graphs}
\label{sec:shapesummary}

Given a set of graph shapes, we are interested in deriving some statistical inferences, 
such as classification, clustering, hypothesis testing, and modeling. The use of a metric 
structure to compute summaries of shapes of graphs is of great importance in 
these analyses. We will use the metric structure introduced earlier to define and compute 
shape statistics -- such as mean, covariance, and PCA -- of given graph data. 

\subsection{Mean Graph Shapes}
Given a set of graph shapes $\{ [A_i] \in {\cal G}, i = 1,2,\dots,m\}$, we define their mean graph shape (Karcher mean) 
to be: 
$$
[A_{\mu}] = \argmin_{[A] \in {\cal G}} \left( \sum_{i=1}^m d_g([A],[A_i])^2 \right)\ .
$$
The algorithm for computing this mean shape is given in Algorithm \ref{algo:mean}. 
\begin{algorithm}
\caption{Graph Mean in ${\cal G}$}
\label{algo:mean}
Given adjacency matrices $A_{i}$, $i=1,..,m$:
\begin{algorithmic}[1]
\State Initialize a mean template $A_{\mu}$.
\State Match $A_{i}$ to $A_{\mu}$ using FGM \cite{zhou2015factorized}
and store the matched graph shape as $A_{i}^*$, for $i = 1,.., m$.
\State Update $A_{\mu} = \frac{1}{m}\sum_{i=1}^{m} A_{i}^*$. 
\State Repeat 2 and 3 until convergence.
\end{algorithmic}
\end{algorithm}

\begin{figure}
\centering
\includegraphics[width=1.1\linewidth]{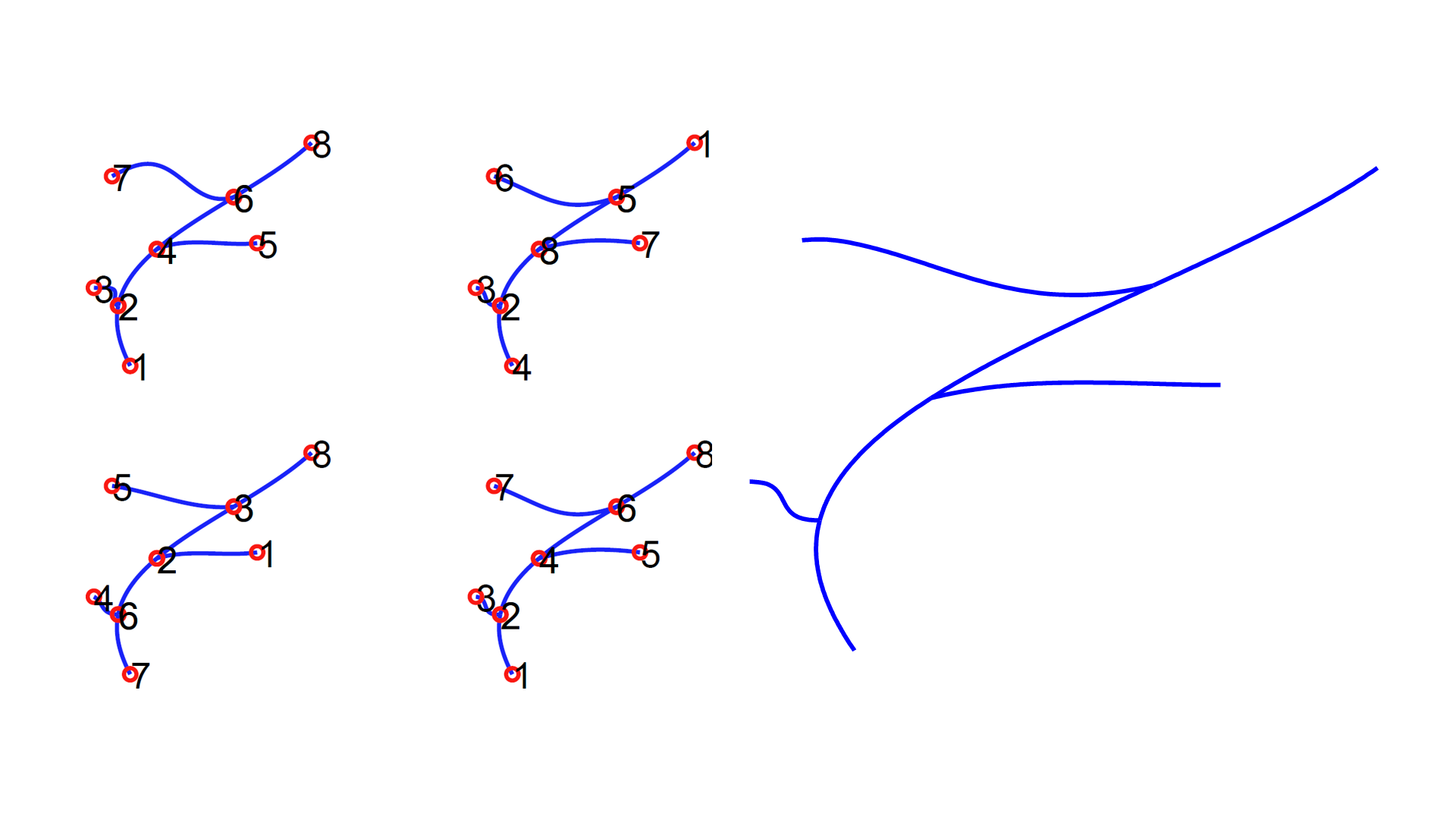}
\caption{Karcher mean of simulated elastic graphs. Left are four sample graph shapes while right is the mean in $\mathcal{G}$}
\label{fig:simu_mean}
\end{figure}

We present an example of computing mean graphs in Fig. \ref{fig:simu_mean}. 
Here we use four simulated tree-like graphs.
They have the same main branch but different side edges and different node ordering.
The simple average does not keep the original structure. 
However, the mean in graph shape space ${\cal G}$ is an appropriate representative of the samples.

\subsection{Tangent Principal Component Analysis (TPCA) in Graph Shape Space}
Graphical shape data is often high dimensional and complex, requiring tools for 
dimension reduction for analysis and modeling. 
In past shape analysis, the tangent PCA has been used for performing 
dimension reduction and for discovering dominant modes of variability in the shape 
data. Given the graph shape metric $d_g$ and the definition of shape mean $A_{\mu}$, we 
can extend TPCA to graphical shapes in a straightforward manner. 
As mentioned earlier, due to the non-registration of nodes in the raw data 
the application of TPCA directly in $\mathcal{A}$ will not be appropriate. 
Instead, one can apply TPCA in the quotient space $\mathcal{G}$, as described in  Algorithm \ref{algo:PCA}. 
After TPCA, graphs can be represented using low-dimensional Euclidean coefficients, which facilitates further 
statistical analysis.

\begin{algorithm}
\caption{Graph TPCA in $\mathcal{G}$}
\label{algo:PCA}
Given adjacency matrices $A_{i}$, $i=1,..,m$:
\begin{algorithmic}[1]
\item Find the mean $A_{\mu}$ using Algorithm \ref{algo:mean} and 
find the matched graph $A_i^*, i = 1,2,..,m$.
\item Get the shooting vector $v_i$ of $A_i^*$ on $T_{A_{\mu}}(\mathcal{G})$ (the tangent space at $A_{\mu}$)  and perform PCA. 
Obtain directions and singular values for the principal components. 
\end{algorithmic}
\end{algorithm}

\begin{figure}
\centering
\includegraphics[width=1\linewidth]{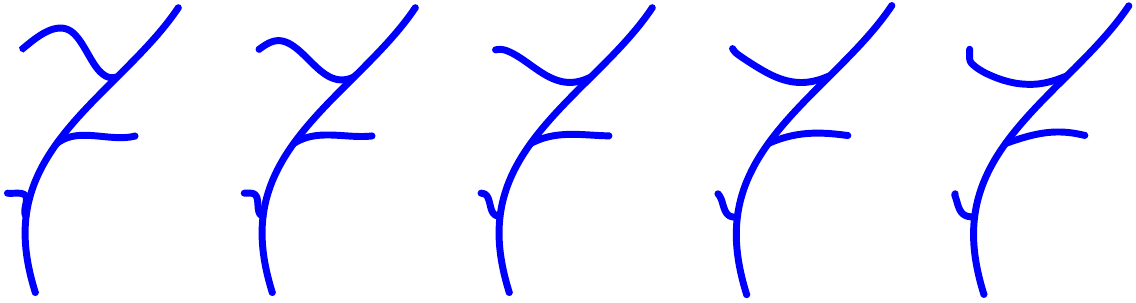}
\caption{Variation of simulated elastic graphs along the first principal direction. The middle one is the mean while the right sides and left  sides are perturbation from mean by $\pm 1,\pm2$ square-root of the first singular value.}
\label{fig:simu_pca}
\end{figure}

An examples of TPCA for graphical shapes is shown in Fig. \ref{fig:simu_pca}. 
The data used here is the same as that in Fig. \ref{fig:simu_mean}.
As we can see from the figure, the first principal variation mainly comes from 
changes in shapes of side edges, since the main edges are essentially the same across all the matched graphs.

\section{Real Data Analysis: Neurons}
\label{sec: neuron}
We first show an application of proposed methods from neuron morphology.
The neurons are from hippocampus of mouse brains~\cite{ascoli2007neuromorpho,revest2009adult}, downloaded from \href{http://neuromorpho.org}{neuromorpho.org}.
Each edge of neurons are curves in $\mathbb{R}^3$.

\begin{figure}
\includegraphics[width=1\linewidth]{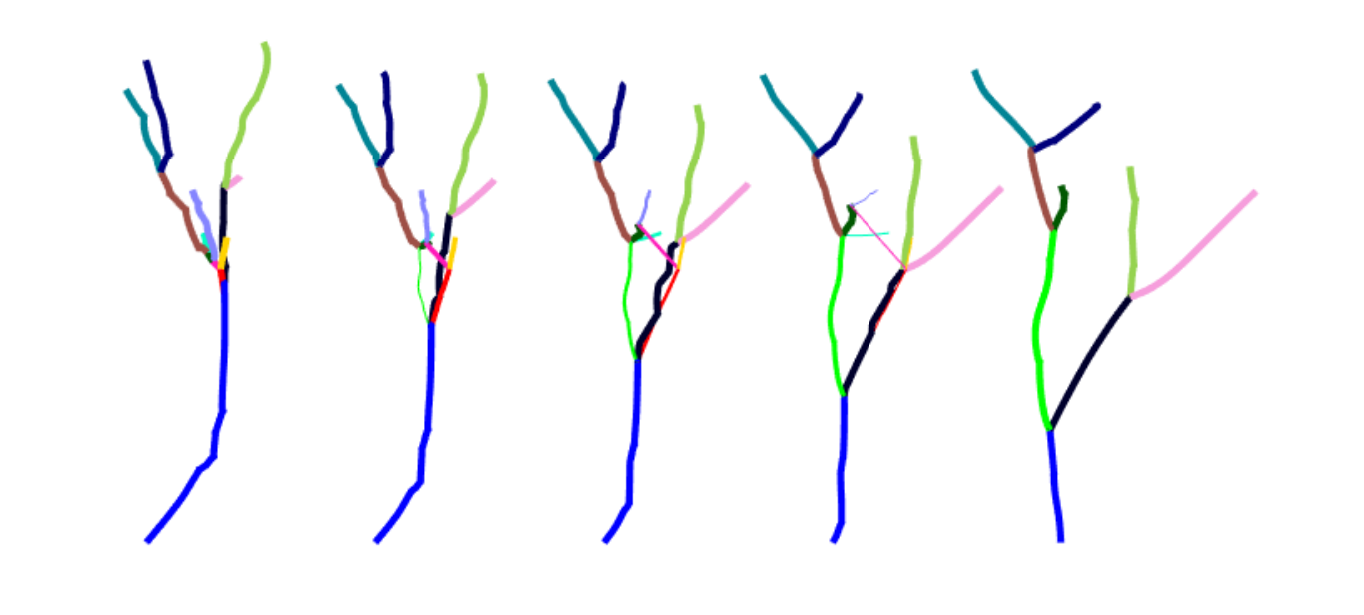}
\caption{Geodesic in $\mathcal{G}$ between two neurons.}
\label{fig:neuron_geodesic}
\end{figure}

Under the proposed metric, Fig. \ref{fig:neuron_geodesic} shows an example of geodesic in $\mathcal{G}$ between two neurons
with different branching structures.
The first one and the last graphs represented the given two neurons, and the 
intermediate shapes show points along the geodesic.
One can see intermediate graphs largely preserve anatomical features across shapes.

\begin{figure}
\includegraphics[width=1\linewidth]{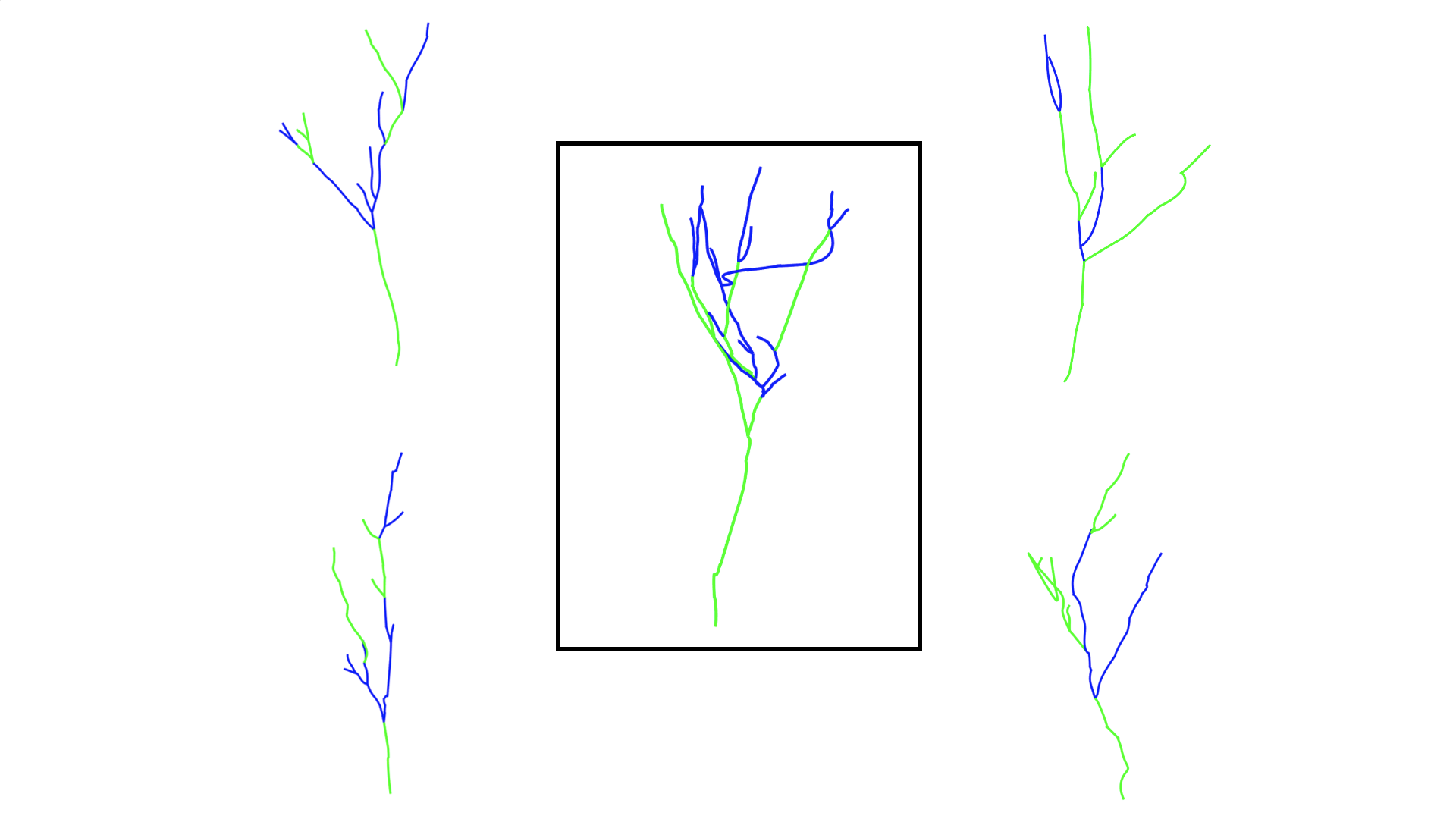}
\caption{Mean shape of 10 neurons, surrounded by four sample neurons. Green color denotes the mostly matched edges.}
\label{fig:neuron_mean}
\end{figure}

\begin{figure}
\includegraphics[width=1\linewidth]{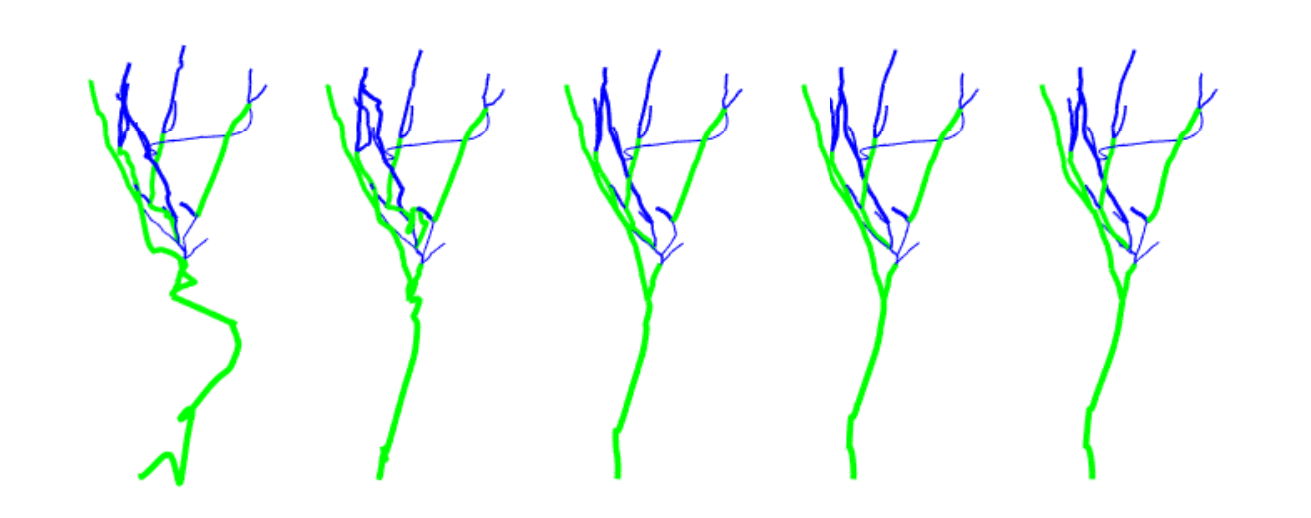}
\caption{Same as Fig. \ref{fig:simu_pca} but for neuron example.}
\label{fig:neuron_pca}
\end{figure}

We also compute the mean shape of $10$ neurons and display it in Fig. \ref{fig:neuron_mean}.
To preserve salient biological structure, we use the largest neuron in the set as a template and match all the other neurons to it, 
when computing the mean.
The middle framed shapes in the picture is the mean neuron, surrounded by four of the ten sample neurons. 
Green color indicates a subset of matched edges.
The mean neuron has relatively smoother edges as compared to sample neurons.
After the graph shapes are registered, we can obtain the principal variations using 
Algorithm \ref{algo:PCA} and we plot the first principal variation of neurons in Fig. \ref{fig:neuron_pca}.
One can see the first principal variation mainly comes from the root edges.

\section{Real Data Analysis: Brain Arterial Networks}
\label{sec: brain artery tree}

In this section, we apply the proposed framework for analyzing the Brain Artery Tree data~\cite{bullitt2005vessel}. 
Here each observation is a 3D geometric object that represents the network of arteries in a human brain.
This object is reconstructed from a 3D Magnetic Resonance Angiography (MRA) images using a tube-tracking vessel segmentation algorithm~\cite{aydin2009principal,aylward2002initialization}.
Because of the complex nature of arterial networks, previous analyses have focused
mainly on some low-dimensional features extracted from the original data. 
For instance, the use of Topological Data Analysis (TDA) on this data can be found in~\cite{bendich2016persistent}.

\begin{figure}
\centering
\includegraphics[width=0.49\linewidth]{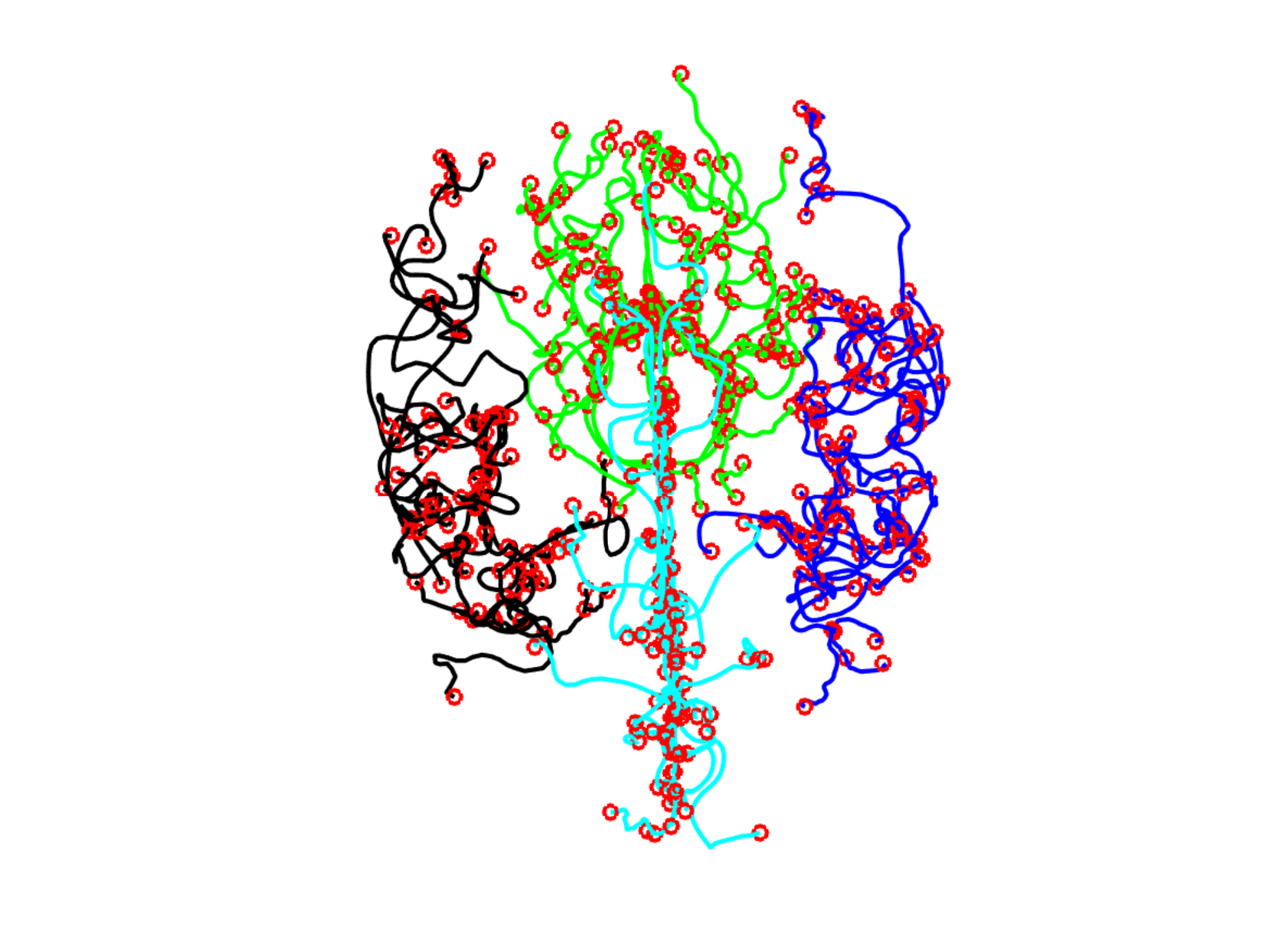}
\includegraphics[width=0.49\linewidth]{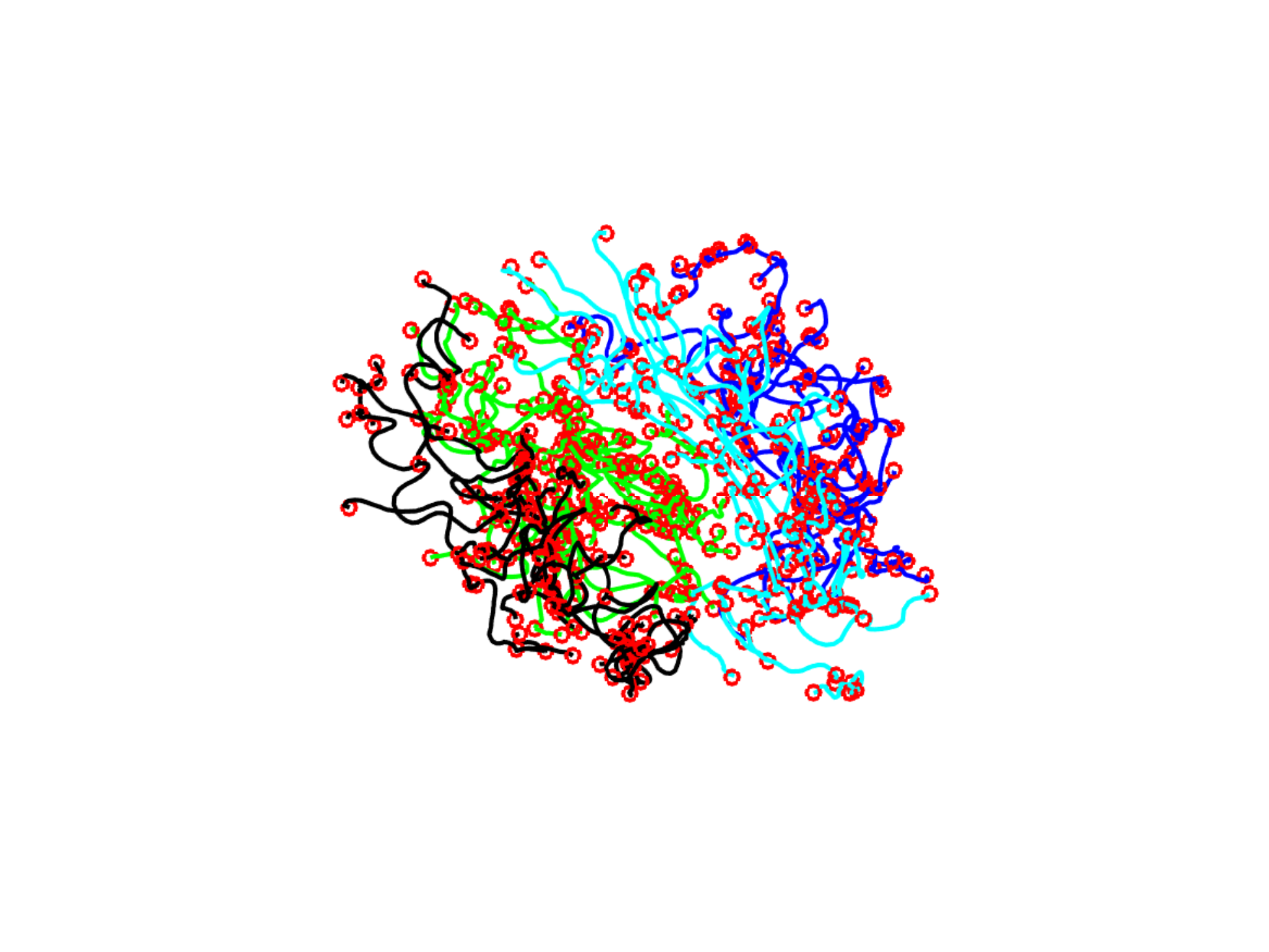}
\caption{An example of Brain Artery Tree data. The four components: 
top, left, bottom, and right are shown in cyan, black, green, and blue, respectively. Nodes are denoted by red circles.}
\label{fig:brain_artery_example}
\end{figure}

\begin{figure}
\centering
\subfloat[Left Components]{
\includegraphics[width=0.53\linewidth]{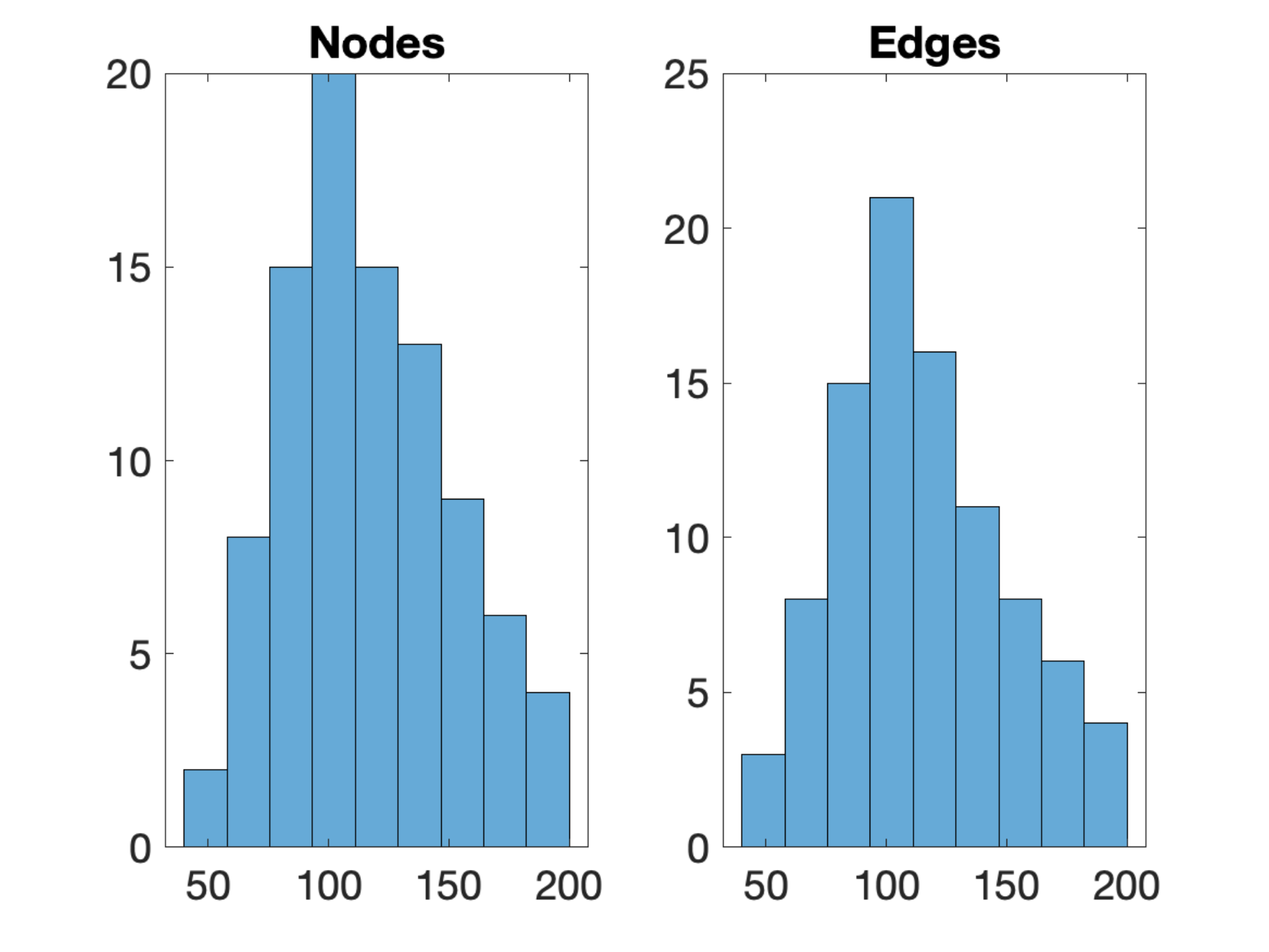}
}
\subfloat[Right Components]{
\includegraphics[width=0.53\linewidth]{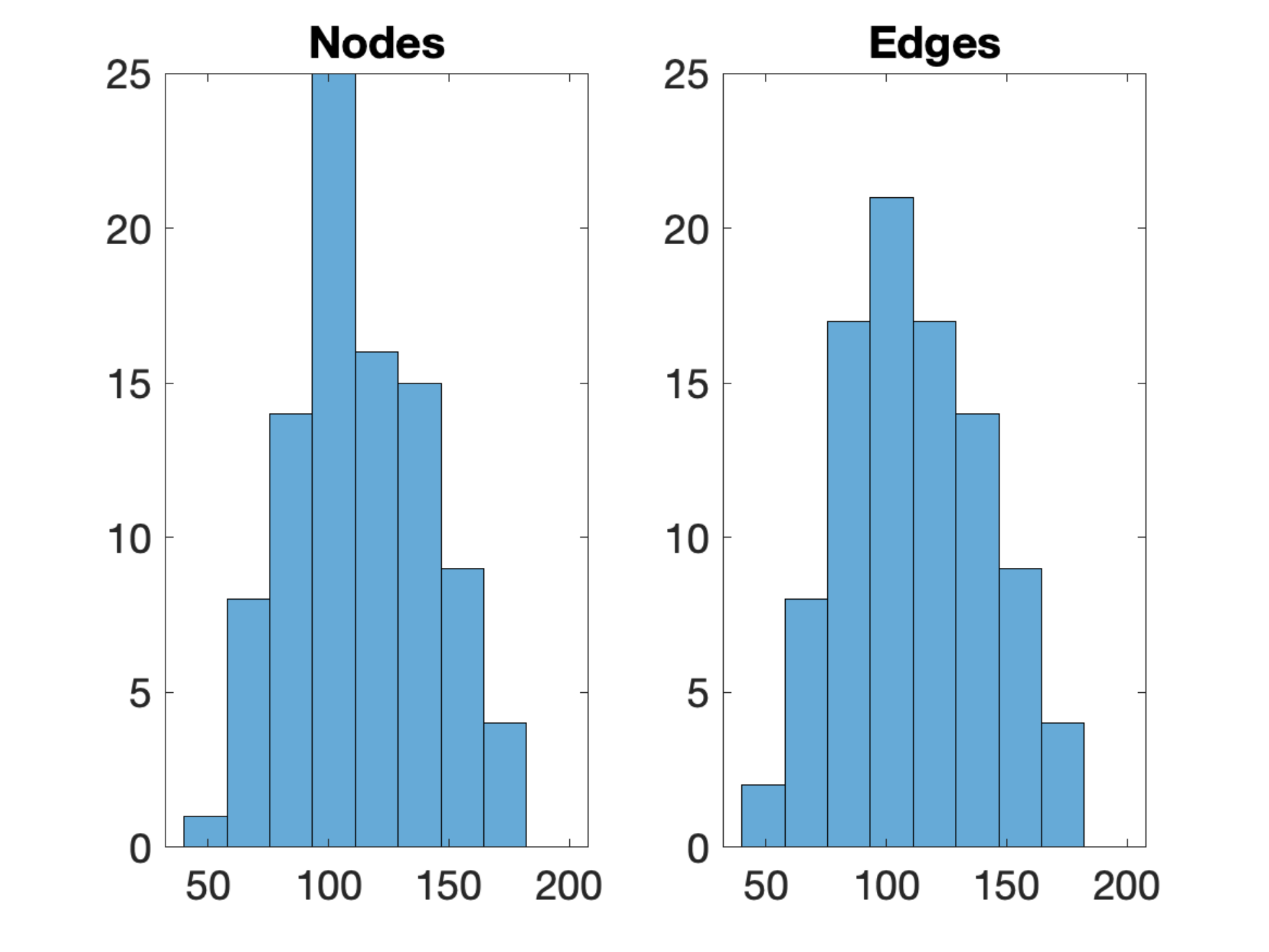}
}
\caption{Histogram of number of nodes and edge in left and right components}
\label{fig:brain_artery_hist}
\end{figure}

We study the data from a geometric point of view and analyze full shapes of these networks.
From an anatomical perspective, it is natural to divide the full network into 
four natural components, as shown in  Fig. \ref{fig:brain_artery_example}. This division helps us focus on comparisons of 
individual components across subjects and also makes the computational tasks more efficient.  
The original data has $98$ subjects but 
we remove six subjects that are difficult to separate into small components.
In the following analysis, we mainly focus on the left and right components with sample size $92$.
In Fig. \ref{fig:brain_artery_hist}, we provide some relevant statistics on the 
numbers of nodes and edges in the left and the right components. 
As these histograms show, the given networks differ drastically in terms of the numbers of nodes and edges across subjects. 
Additionally, there are large differences in both the shapes and the patterns of arteries forming these networks. 
Consequently, the problem of analyzing shapes of these arterial networks is quite challenging, and has not 
been done in a comprehensive way in the past. 
\\

\begin{figure}
\includegraphics[width=1\linewidth]{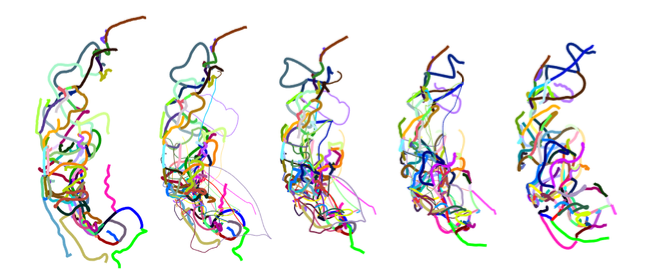}
\includegraphics[width=1\linewidth]{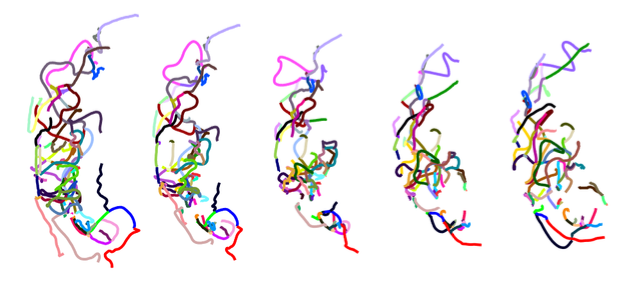}
\caption{Geodesic between two left components. Top row is the original graph geodesic in $\mathcal{G}$. For better visualization, we throw away the unmatched edges and show the geodesic again in the bottom.}
\label{fig:brain_artery_left_geodesic}
\end{figure}

\noindent {\bf Geodesic Deformations}: 
As a first step, we use the techniques developed in this paper to compute geodesic paths between 
arbitrary arterial networks. 
We show an example of geodesic between two left components in Fig. \ref{fig:brain_artery_left_geodesic}. 
In the top row, the first and the last graphs are the two left components of brain arteries. 
The intermediate graphs represent the optimal deformation from one to another along 
a geodesic in $\mathcal{G}$.
To improve visual clarity we remove some unmatched edges from the graphs
and plot the same geodesic again in the bottom row.
One can use the color to track the deformation of each edge. These geodesics 
are useful in several ways. They provide registrations of arteries across networks and they
help follow deformations of matched arteries from one network to another. 
\\

\begin{figure}
\begin{center}
\includegraphics[width=1\linewidth]{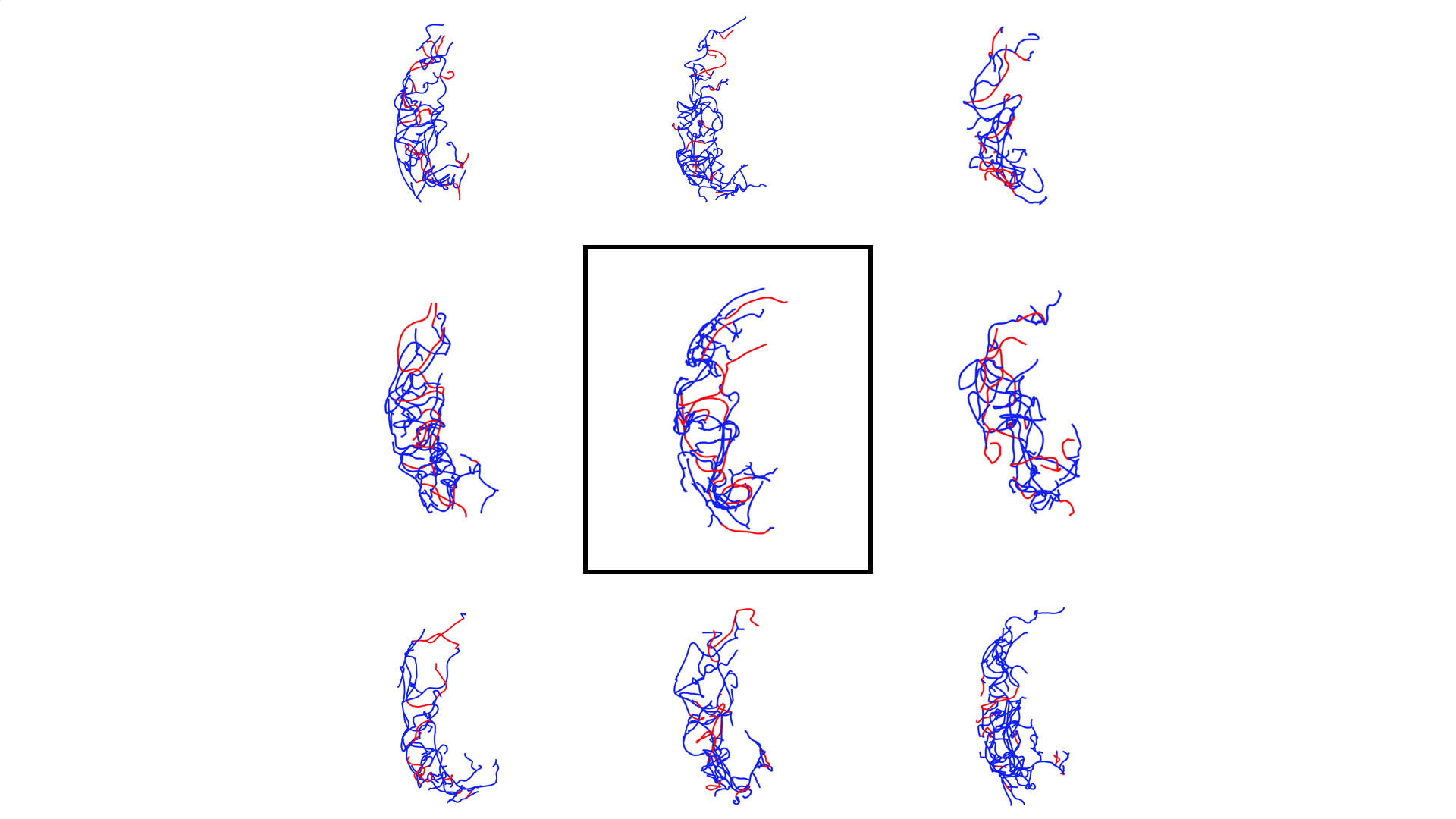}
\caption{Mean of left components of brain artery tree data. Middle one is the mean of 92 subjects while the surroundings are eight examples. Red edges denotes mostly matched edges.}
\label{fig:brain_artery_mean-left}
\end{center}
\end{figure}

\begin{figure}
\begin{center}
\includegraphics[width=1\linewidth]{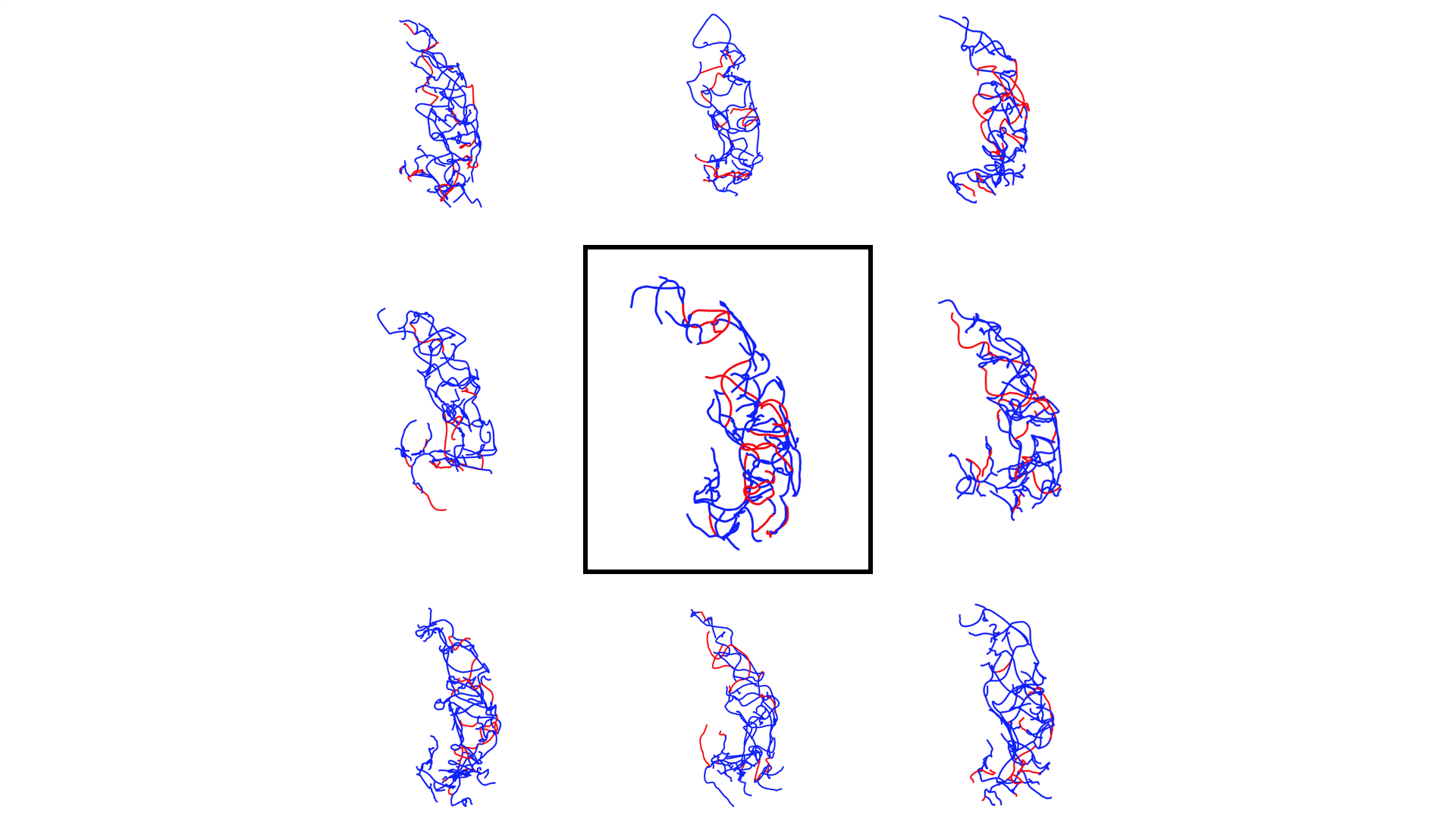}
\caption{Mean of right components of brain artery tree data. Middle one is the mean of 92 subjects while the surroundings are eight examples. Red edges denotes mostly matched edges.}
\label{fig:brain_artery_mean-right}
\end{center}
\end{figure}

\noindent {\bf Mean Arterial Networks}:
Given $92$ sample arterial networks, it is interesting and useful to be able to compute their mean shape.
To accelerate computation, we approximated and simplified the process as follows. 
We basically registered each graph shape to the largest size graph in the dataset and 
used that fixed registration to compute the mean. This is not quite the optimal registration prescribed in 
the mean algorithm but provides a decent approximation. 
Figs. \ref{fig:brain_artery_mean-left} and \ref{fig:brain_artery_mean-right} show the mean shapes for the left and the right components, 
respectively.
In both figures, the middle framed shape is the mean graph, surrounded by eight of 92 samples used in the 
computations.
We use red color to denote some subset of matches edges in the individual networks and the mean.
The mean shapes show a smoother representation of individual graph shapes and largely preserve 
network patterns present in the data.
\\

\noindent {\bf Effects of Covariates on Shapes}: 
An important use of shape analysis of artery data is in studying the 
the effects of covariates such as gender and age on the brain arteries.  
Due to the high dimensionality and complex nature of brain arterial network, we 
first perform Graph PCA using 
Algorithm \ref{algo:PCA} to project each brain arteries into a low-dimensional vector space.  
An arterial network with 197 nodes and 50 discrete points in each edge has a dimension of $3 \times 50 \times 197 \times 197 = 5821350$.
However, by using Graph PCA, We use approx. 60 principal components to obtain over $80 \% $ variation
in the original data.
In order to avoid the confounding effect of artery size, we rescale edges by the total artery length. 
As the result, we can focus on gender and age effects on the shapes of brain arterial networks.

To study the effect of gender on the arterial graph shapes, 
we implement a two sample t-test on the first principal scores and a Hotelling's T-squared test on the first several principal scores. 
The results can be found in Table \ref{tab:brain_artery_gender}.
Although two of the $p$-values are somewhat low ($0.0161$ and $0.0419$),
most of the $p$-values are high. 
(We also applied a permutation test described in the following paragraphs and obtained similar results.)
Thus, we do not find any significant difference between shapes 
of artery networks in female and male brains. 
The result is consistent with 
conclusions in \cite{bendich2016persistent} and~\cite{shen2014functional} although our approach is 
very different from those papers. (We note that
these two papers reported multiple different $p$-values using different methods.)
Whether there is an anatomical shape difference between brain arterial networks of female and male 
subjects remains an open question.

\begin{table}
\centering
\begin{tabular}{|c|ccccc|} 
 \hline
         & t test & \multicolumn{4}{c|}{Hotelling's T-squared test }  \\ 
         & &  2  & 3 & 4 & 5  \\
 \hline
 \hline
 Left  & 0.0161 & 0.3528  & 0.2718 & 0.3949 & 0.4553\\
 \hline
 Right & 0.9866 & 0.2288  & 0.0419 & 0.0830 & 0.1206  \\
 \hline
\end{tabular}
\caption{Hypothesis testing of gender effect on principal scores of shapes of brain arterial networks.}
\label{tab:brain_artery_gender}
\end{table}

\begin{figure}
\subfloat[Left Component]
{
\includegraphics[width=0.5\linewidth]{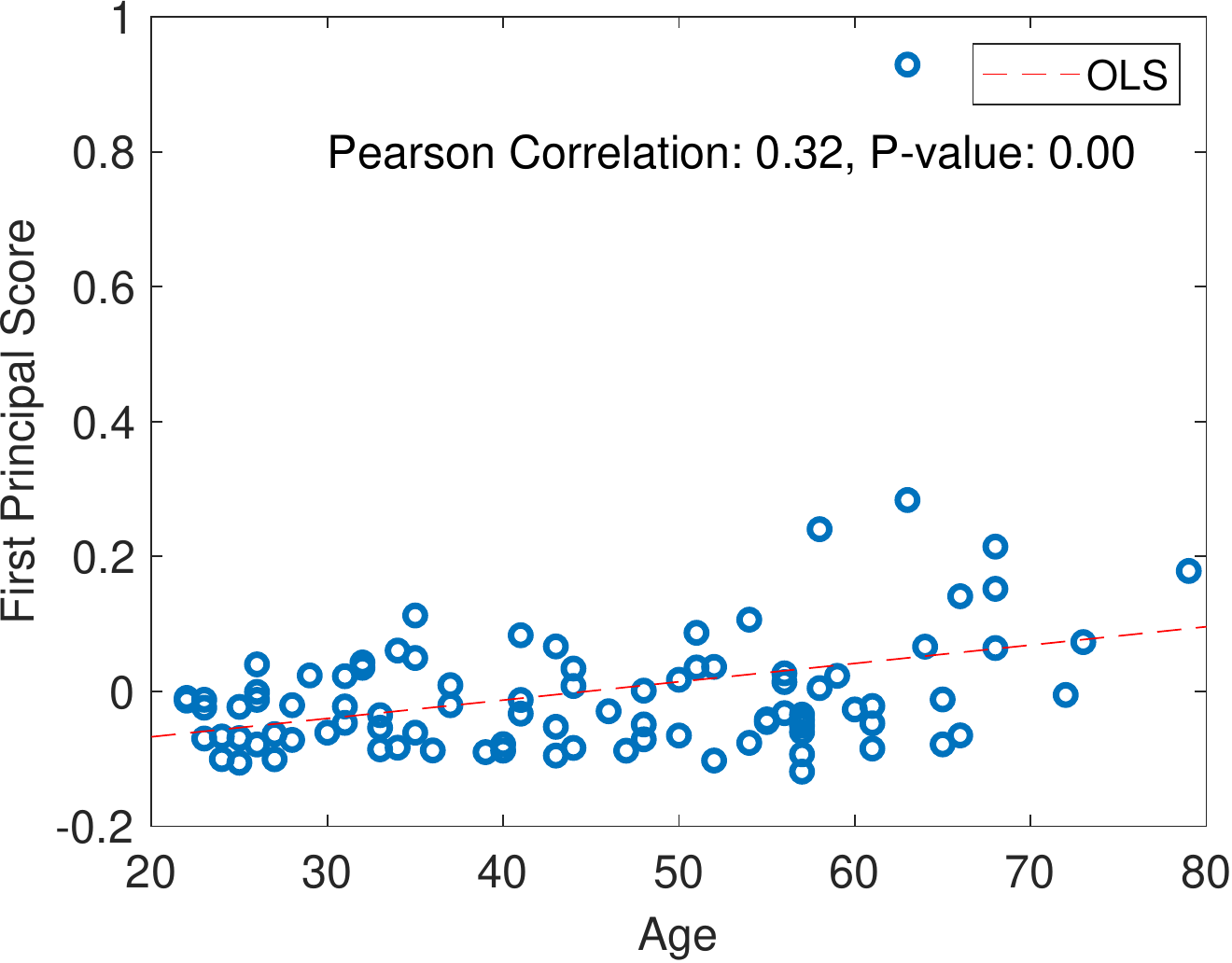}
}
\subfloat[Right Component]
{
\includegraphics[width=0.5\linewidth]{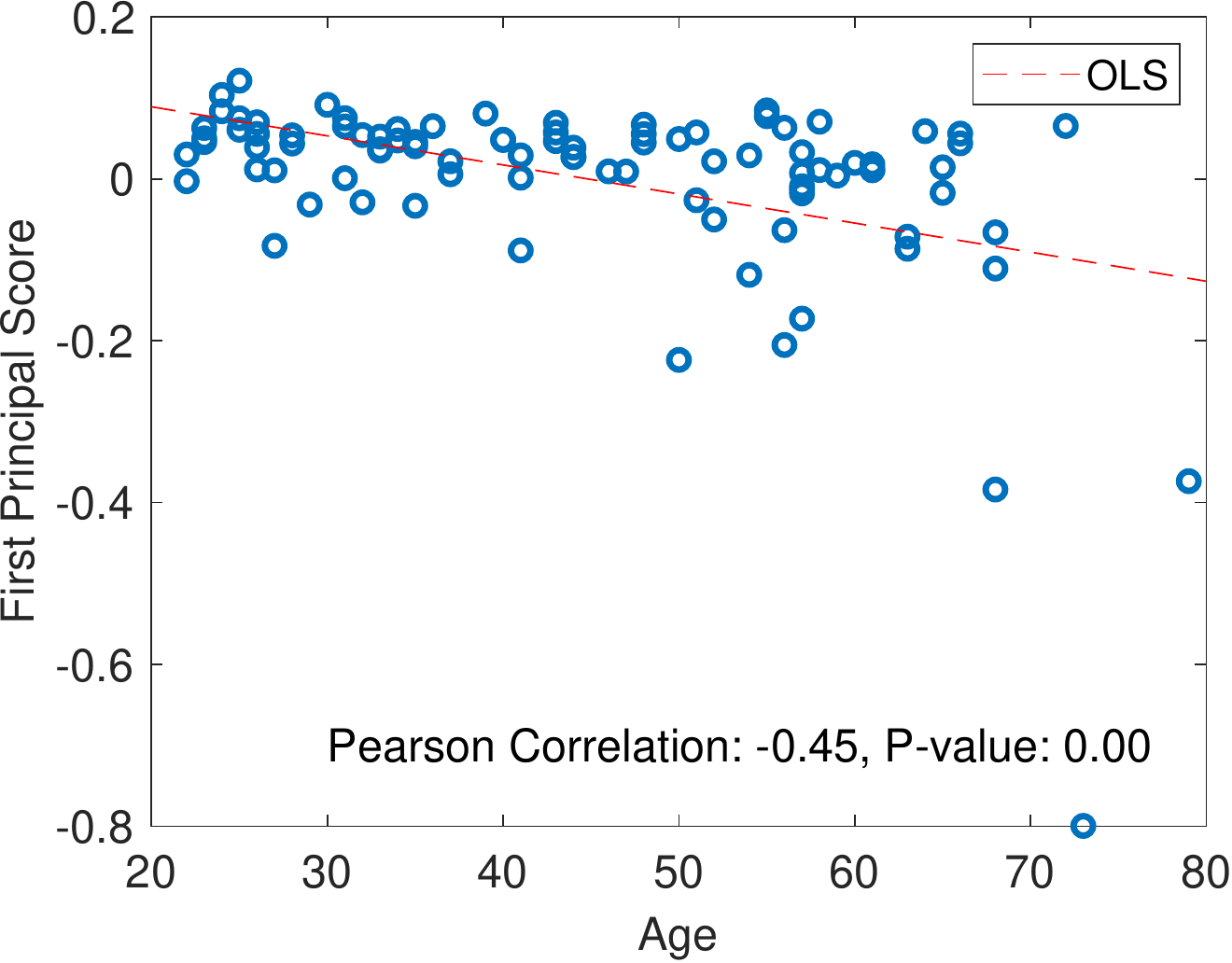}
}
\caption{High correlation between ages and first principal scores of shape of arterial networks.}
\label{fig:brain_artery_age}
\end{figure}

To study the effect of age on these shape, we studied correlations between the
age and the PCA scores of arterial shapes, and the results are shown in Fig. \ref{fig:brain_artery_age}. 
We found a strong linear correlation between age and first principal scores of brain arteries. 
The correlation coefficients between left, right components and ages are $0.32$ and $-0.45$, respectively.
Both of them are significant with $p$-values almost 0.
This result is similar to some published results in the literature but obtained using 
different mathematical representations than ours~\cite{bendich2016persistent,shen2014functional}.
As mentioned before, we also use a permutation test to validate the age effects.

\begin{figure}
\subfloat[Distance matrix of left components]
{
\includegraphics[width=0.52\linewidth]{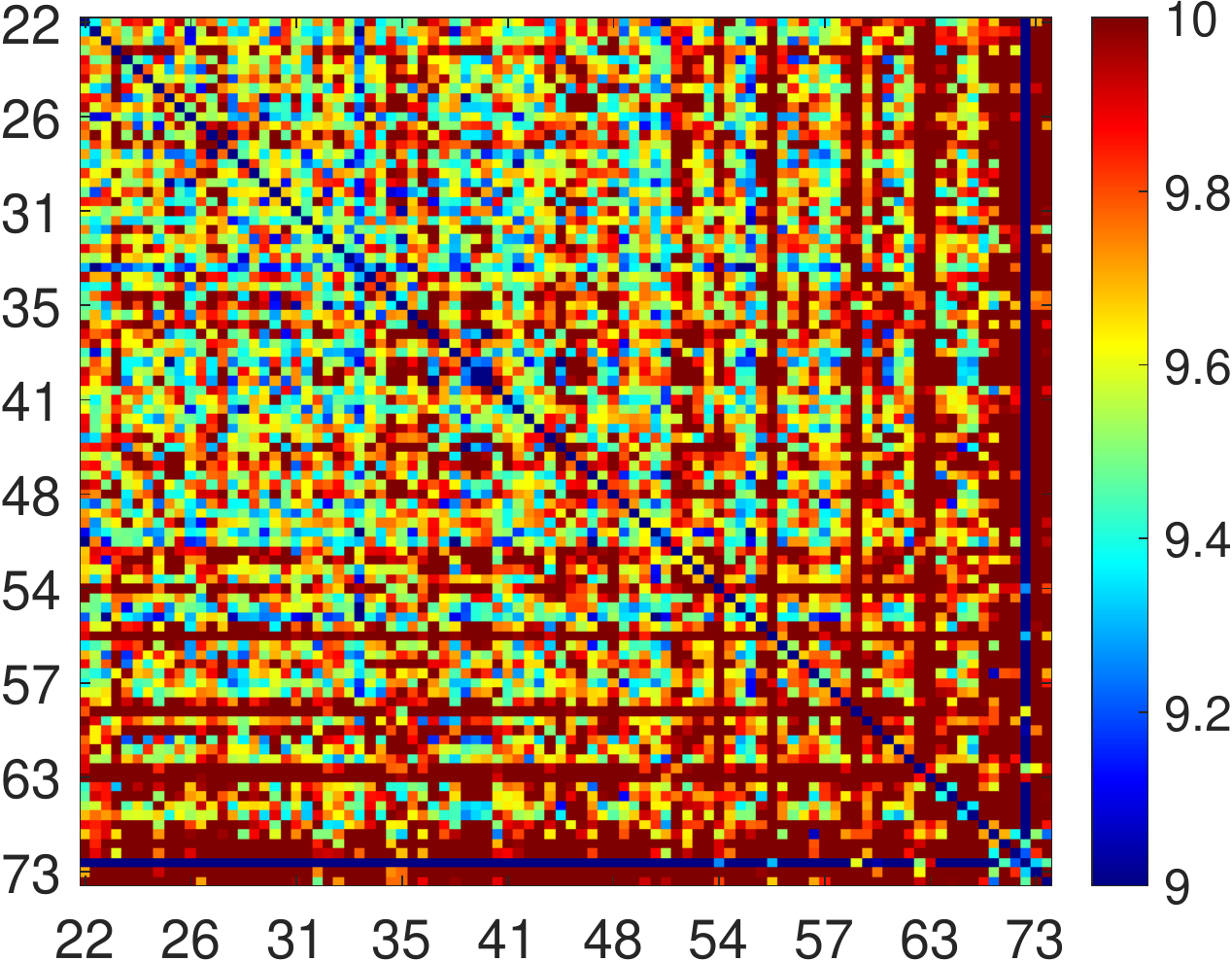}
}
\subfloat[Distance matrix of right components]
{
\includegraphics[width=0.52\linewidth]{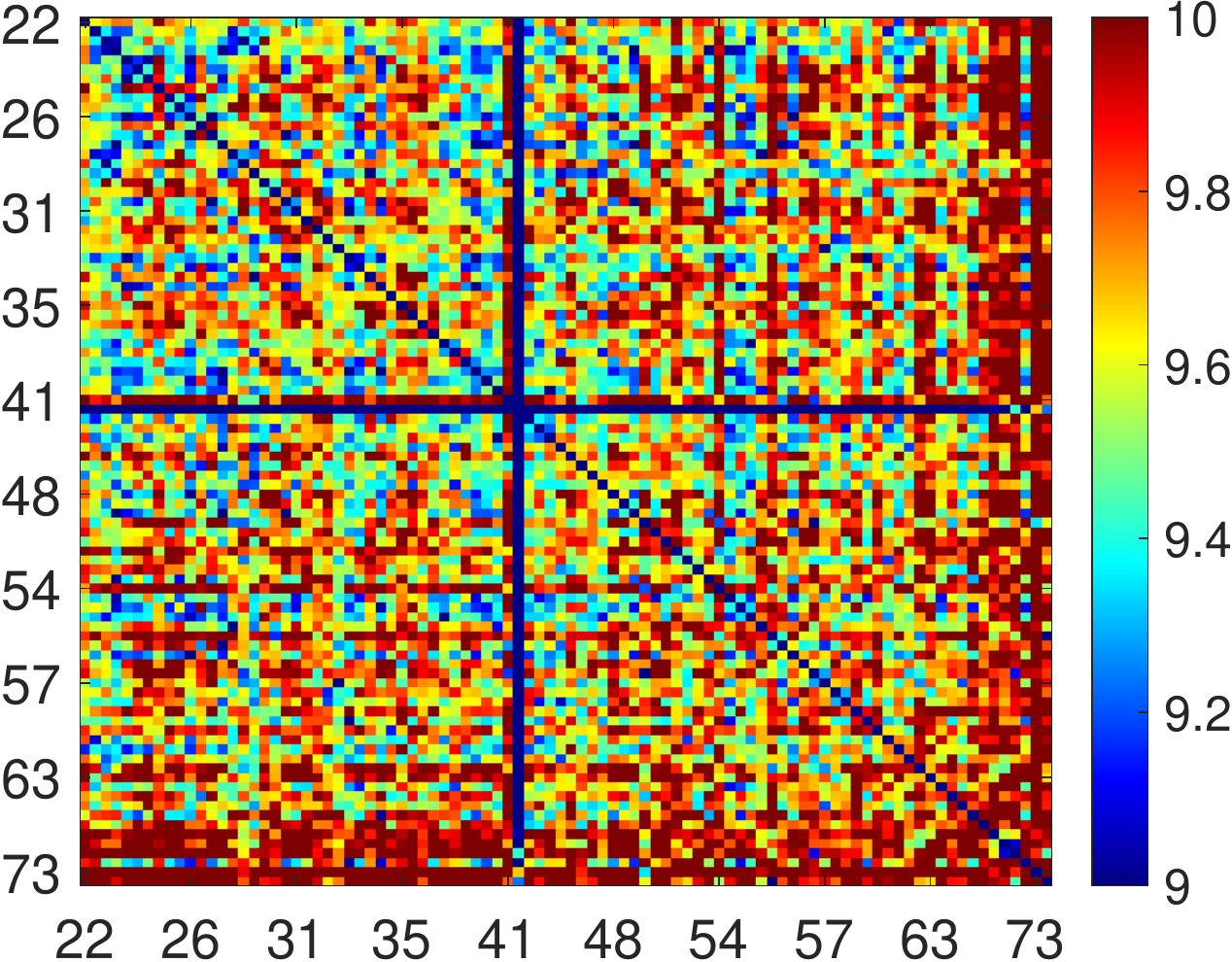}
}\\
\subfloat[MDS plot of left components]
{
\includegraphics[width=0.5\linewidth]{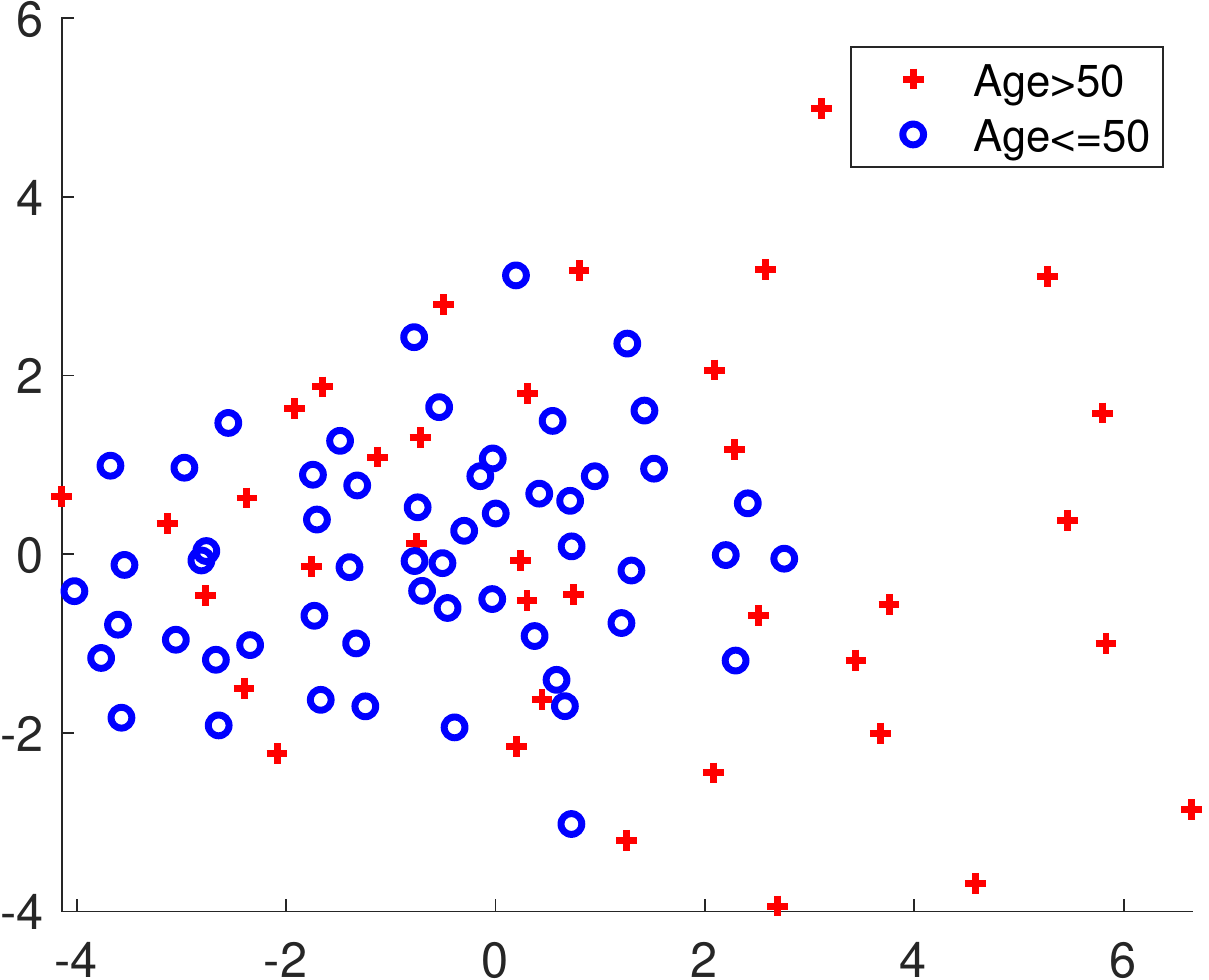}
}
\subfloat[MDS plot of right components]
{
\includegraphics[width=0.5\linewidth]{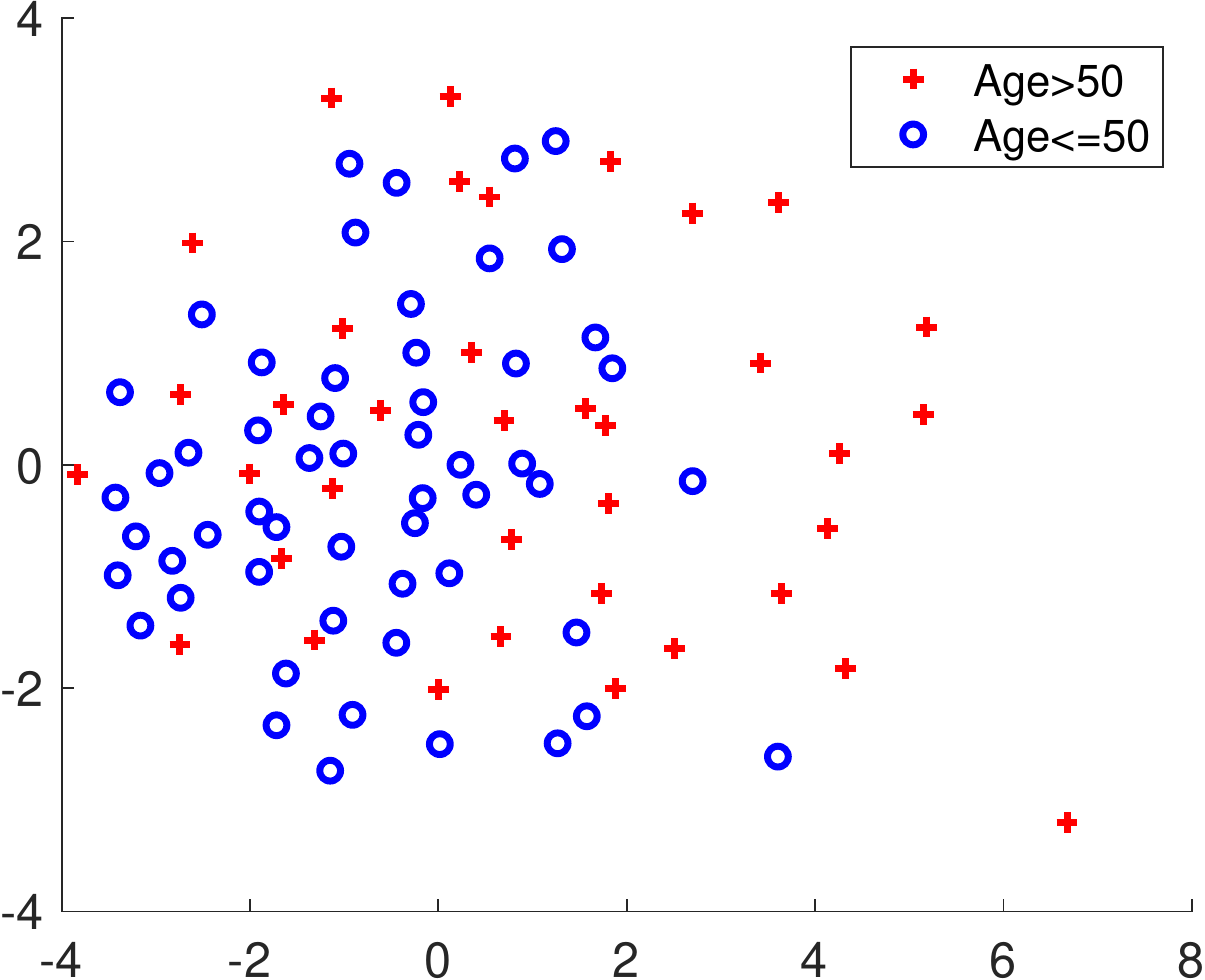}
}
\caption{Distance matrices and MDS plot. The axes of distance matrices are labeled by ages.}
\label{fig:brain_artery_distmat}
\end{figure}

We also investigate the effects of covariates arterial shapes using the shape metric directly. 
Fig. \ref{fig:brain_artery_distmat} shows pairwise distance matrices for left and right components, using the 
shape distance defined in Eqn. \ref{eq:metric}. 
(As mentioned before, we have scaled the edges by the total artery length and thus the distances quantify only shape differences.)
We order the distance matrices by the ages of subjects. The color pattern of pixels in these matrices show that shape distances
increase with the age (darker red colors are towards bottom right). 
This implies that the shape variability in brain arterial networks grows with the age!
Another way to visualize the age effect is by projecting the distance matrix into a
two-dimensional space using Multidimensional scaling (MDS)~\cite{kruskal1964multidimensional}. 
Now it is clear to see the separation between shapes before and after the age 50.

\begin{figure}
\subfloat[Gender effect on left components]
{
\includegraphics[width=0.5\linewidth]{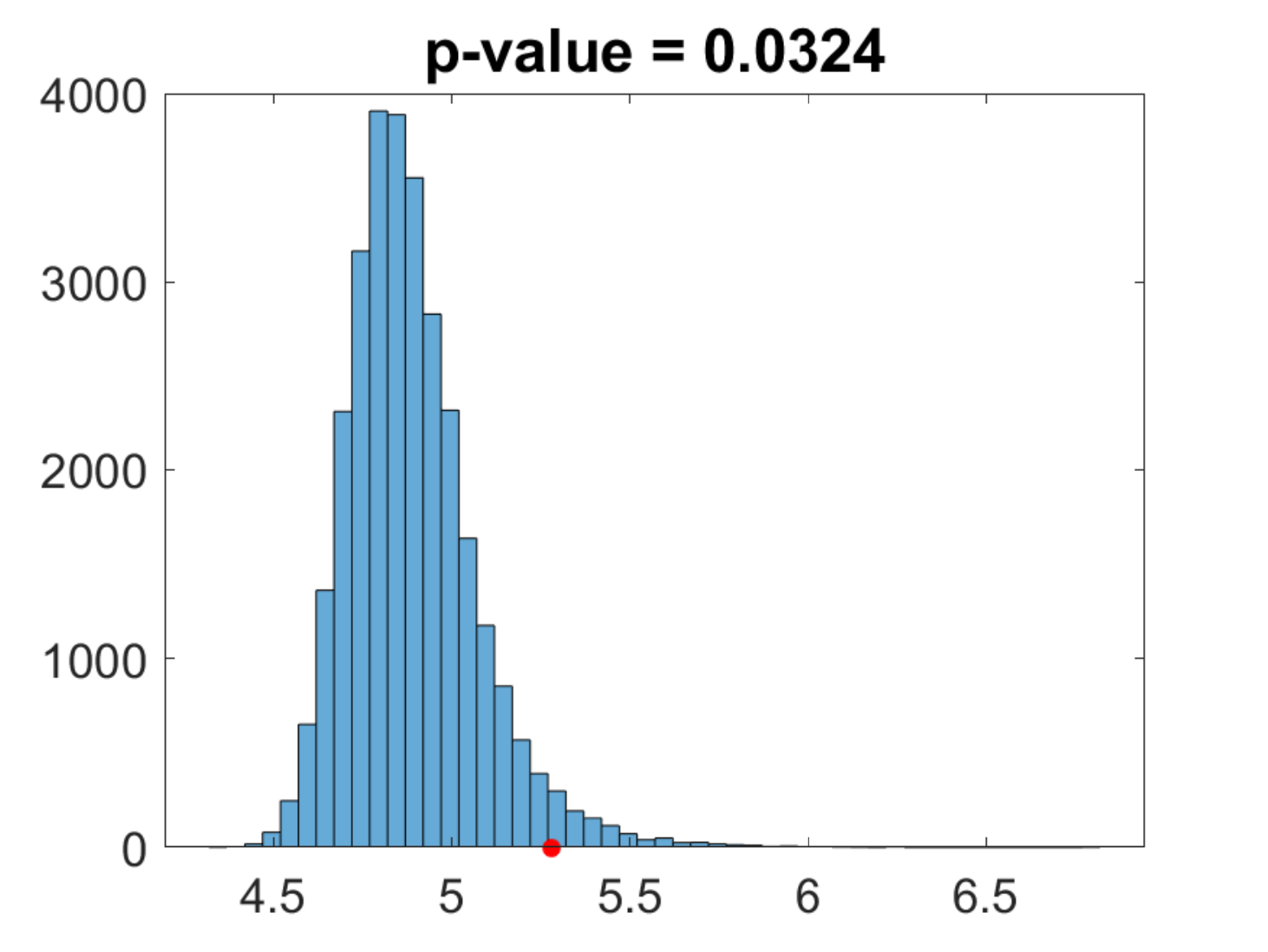}
}
\subfloat[Gender effect on left components]
{
\includegraphics[width=0.5\linewidth]{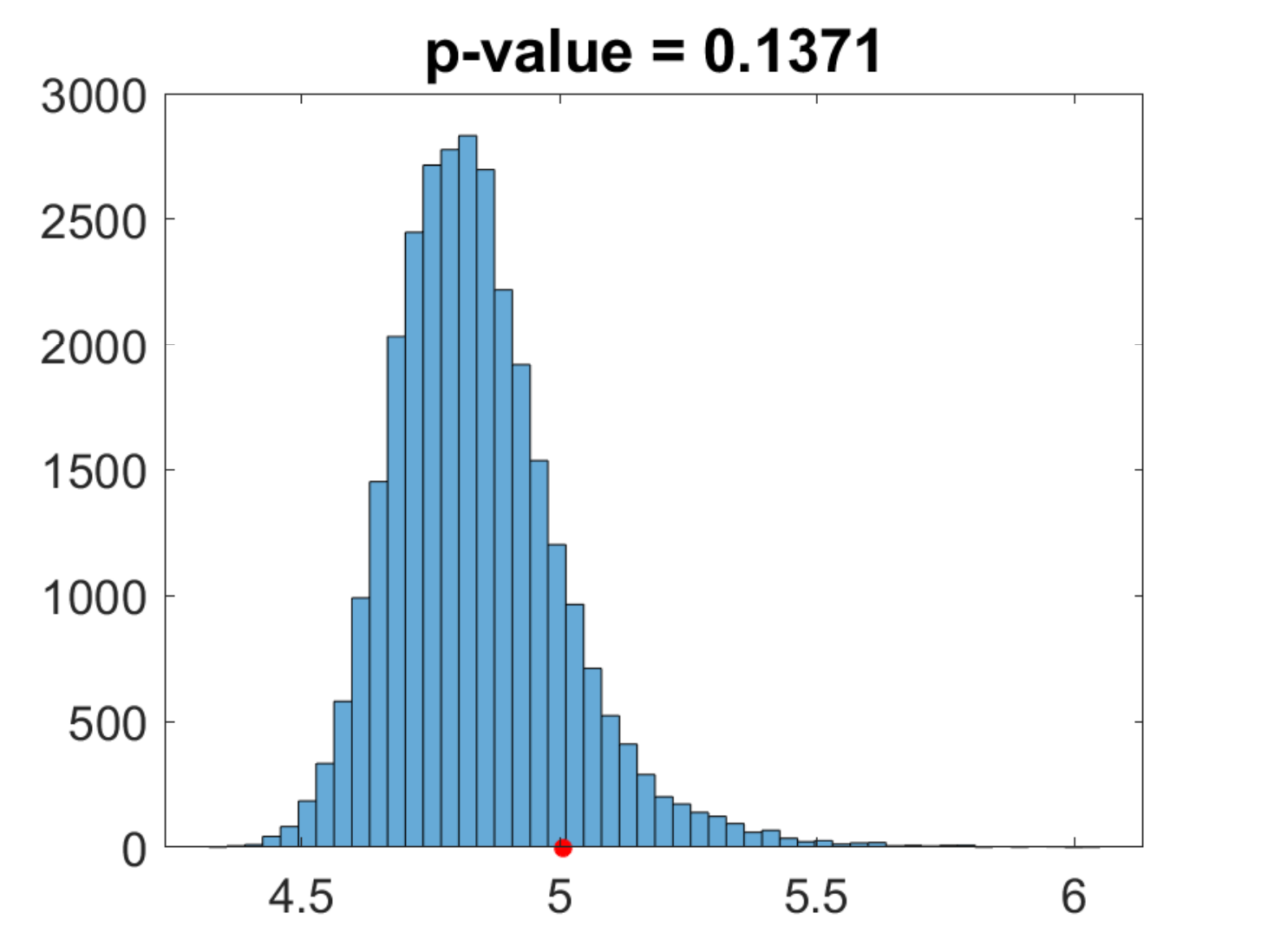}
}\\
\subfloat[Age effect on left components]
{
\includegraphics[width=0.5\linewidth]{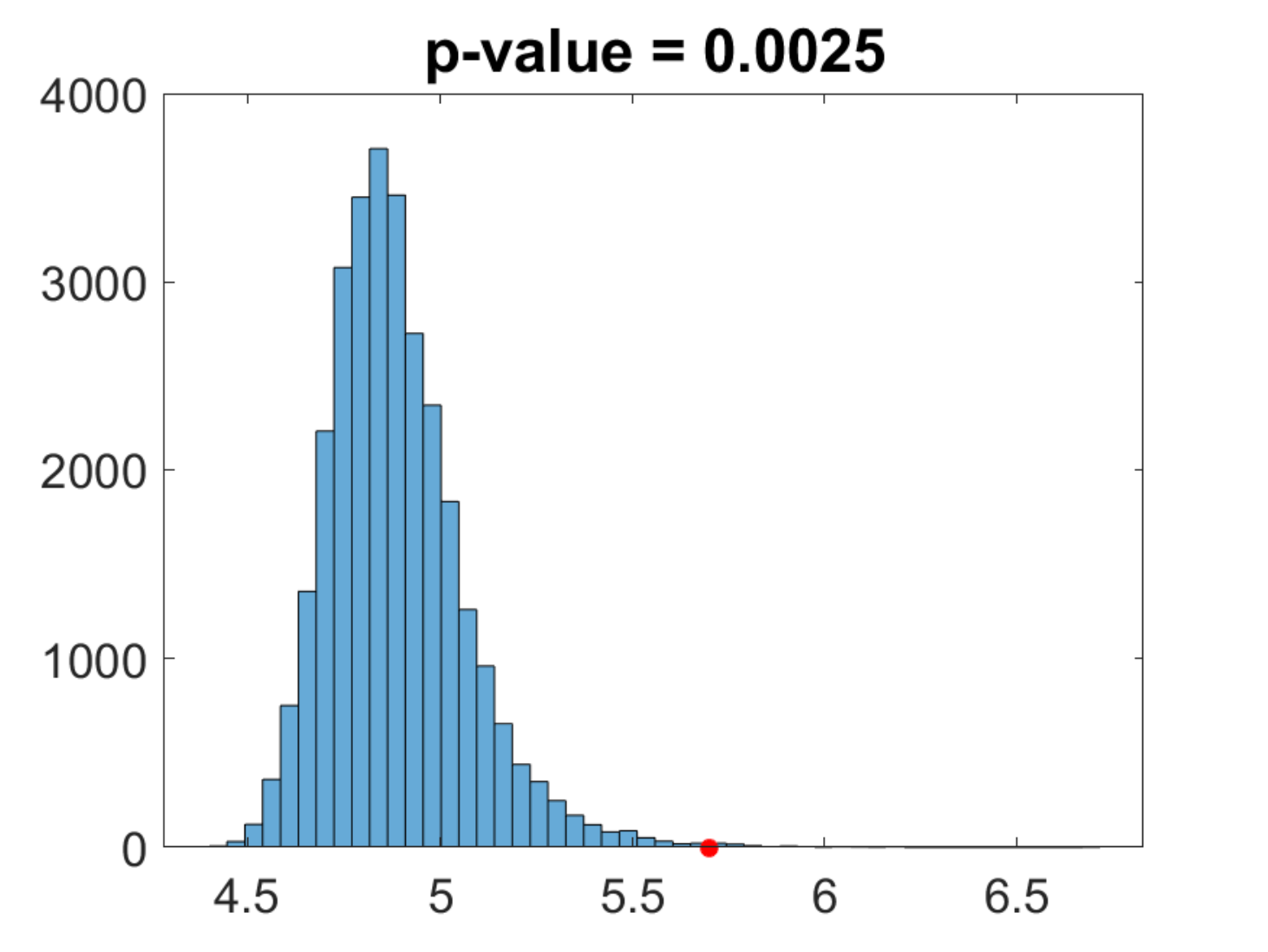}
}
\subfloat[Age effect on right components]
{
\includegraphics[width=0.5\linewidth]{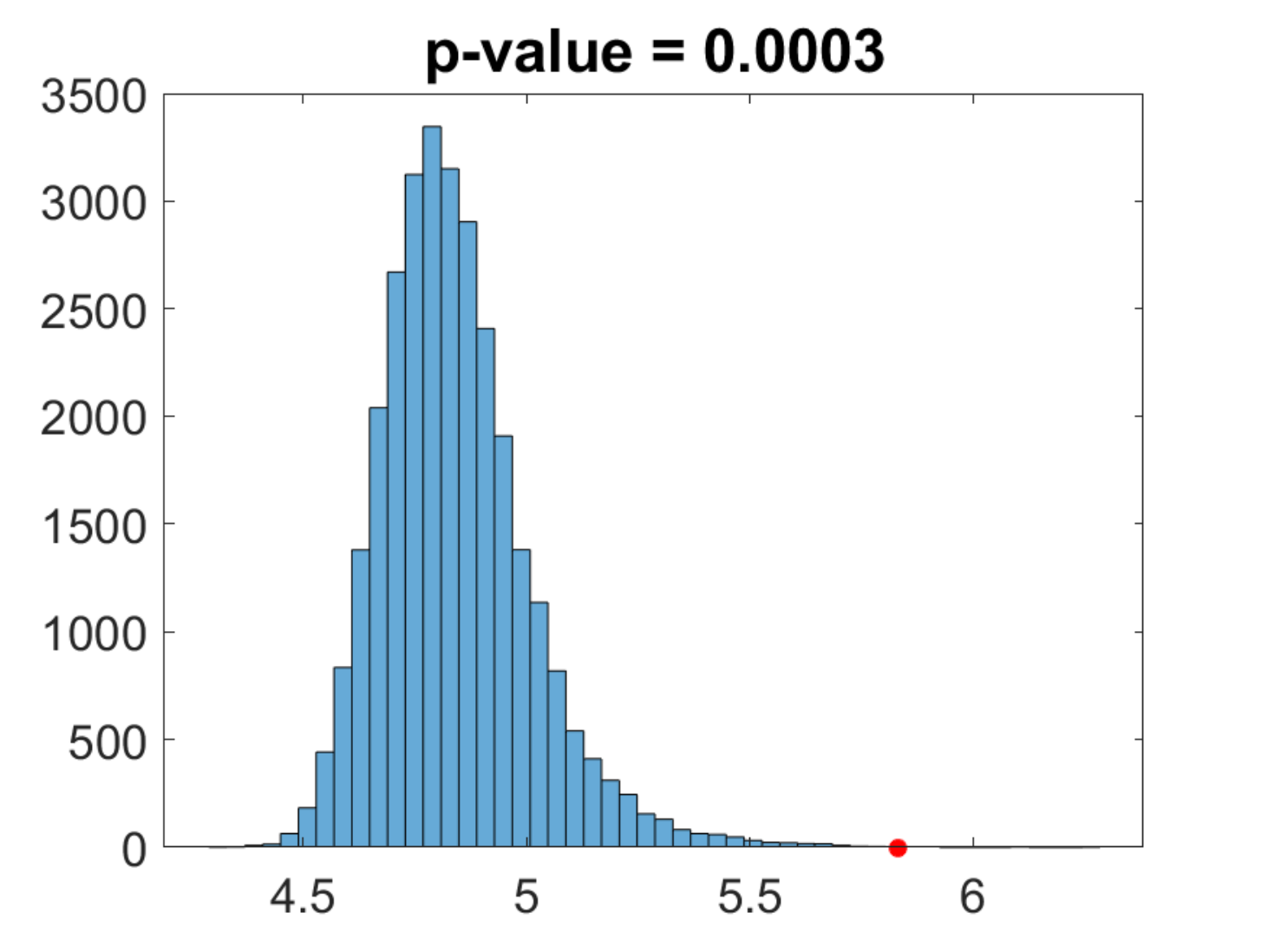}
}
\caption{Permutation test of gender and age effects. Right dots represent the test statistics of current sample.}
\label{fig:brain_artery_anova}
\end{figure}

To further validate the gender and age effects, we implement a permutation test \cite{hagwood-etal:2013} based on shape distances.
The basic idea is, for each time, we randomly assign the brain arteries into different groups (male and female, older than 50 and younger than 50) and compute the test statistics. 
We repeat this process $30000$ times and see what is the $p$-value for the current test statistics.
The result can be found in Fig. \ref{fig:brain_artery_anova}.
While the gender effect still remains unclear, one can see a significant age effect on the brain arteries. 
Note the pairwise distance is compromised because of computational issue. 
Here we first match each graph to the largest graph in the dataset and compute pairwise shape distances without any 
further matching. 

\section{Conclusion }
\label{sec:conclusion}

This paper introduces a comprehensive geometrical framework for analyzing shapes of complex networks and graphs.
These graphs are characterized by edges that are shapes of Euclidean curves and shape analysis involves registration of 
nodes/edges across graphs. This framework can handle graphs that exhibit both topological variability 
(different number and connectivities of edges) and geometric variability (different shapes of edges). The developed framework 
provides tools for quantifying shape differences, computing average shapes, discovering principle modes, and testing 
covariates. These tools are illustrated on complex objects such as neurons and brain arterial networks. Future research 
in this area involves improvement of optimization tools for graph matching, and studying regression models in data involving 
human brain networks.

\appendix
\section{Elastic shape analysis of curves}
The edges in elastic graphs are Euclidean curves and to analyze their shapes 
we use elastic shape analysis framework described in \cite{srivastava2016functional}
and several other places.
Let $\beta(t): [0,1] \rightarrow \mathbb{R}^n, n=2,3$ represent a parametrized curve.
In this paper, $\beta$ represents an edge connecting two nodes of an elastic graphs.
Define the square root velocity function (SRVF) of $\beta$ as:
$q(t)=
\frac{\dot{\beta}(t)}{\sqrt{|\dot{\beta}(t)|}}$, if $|\dot{\beta}(t)| \neq 0$ and zero otherwise. 
One can recover $\beta$ from its SRVF using $\beta(t) = \beta(0) + \int_{0}^t q(s)|q(s)| ds$.
If $\beta$ is absolutely continuous, the SRVF is square-integrable, i.e., $q \in \mathbb{L}^2$.
It can be shown the $\mathbb{L}^2$ norm on SRVF space is an elastic Riemannian metric on the original curve space.
Therefore, one can compute the elastic distance between two curves $\beta_1, \beta_2$ using  
$d(\beta_1,\beta_2) = \|q_1-q_2 \|_{\mathbb{L}^2}$.

One of the most import challenges in shape analysis is registration issue, i.e., finding the point correspondence between curves. 
Let $\gamma : [0,1] \rightarrow [0,1]$ represent a boundary-preserving diffeomorphism.
The action of diffeomorphism group on an SRVF $q$ is $q*\gamma = (q \circ \gamma)\sqrt{\dot{\gamma}}$.
Next, one mods out the shape-preserving group actions: rotation and re-parametrization as follows.
Each shape can be represented by orbits formed by rotation and re-parametrization group: $[q] = \{O(q*\gamma)|O\in SO(n), \gamma \in \Gamma\}$. 
The space of $[q]$ is shape space $\mathcal{S}$.
The metric for shape space is: 
$d_s([q_1],[q_2]) = \inf_{\gamma,O} \|q_1 - O(q_2*\gamma)\|$. One can use this metric to define and compute
averages of shapes of curves and their PCA analysis as described in \cite{srivastava2016functional}.

\cleardoublepage

{\small
\bibliographystyle{ieee_fullname}
\bibliography{egbib}

\begin{thebibliography}{10}\itemsep=-1pt

\bibitem{ascoli2007neuromorpho}
Giorgio~A Ascoli, Duncan~E Donohue, and Maryam Halavi.
\newblock Neuromorpho. org: a central resource for neuronal morphologies.
\newblock {\em Journal of Neuroscience}, 27(35):9247--9251, 2007.

\bibitem{aydin2009principal}
Burcu Ayd{\i}n, G{\'a}bor Pataki, Haonan Wang, Elizabeth Bullitt, James~Stephen
  Marron, et~al.
\newblock A principal component analysis for trees.
\newblock {\em The Annals of Applied Statistics}, 3(4):1597--1615, 2009.

\bibitem{aylward2002initialization}
Stephen~R Aylward and Elizabeth Bullitt.
\newblock Initialization, noise, singularities, and scale in height ridge
  traversal for tubular object centerline extraction.
\newblock {\em IEEE transactions on medical imaging}, 21(2):61--75, 2002.

\bibitem{bendich2016persistent}
Paul Bendich, James~S Marron, Ezra Miller, Alex Pieloch, and Sean Skwerer.
\newblock Persistent homology analysis of brain artery trees.
\newblock {\em The annals of applied statistics}, 10(1):198, 2016.

\bibitem{bubenik2015statistical}
Peter Bubenik.
\newblock Statistical topological data analysis using persistence landscapes.
\newblock {\em The Journal of Machine Learning Research}, 16(1):77--102, 2015.

\bibitem{bullitt2005vessel}
Elizabeth Bullitt, Donglin Zeng, Guido Gerig, Stephen Aylward, Sarang Joshi,
  J~Keith Smith, Weili Lin, and Matthew~G Ewend.
\newblock Vessel tortuosity and brain tumor malignancy: a blinded study1.
\newblock {\em Academic radiology}, 12(10):1232--1240, 2005.

\bibitem{caelli2004eigenspace}
Terry Caelli and Serhiy Kosinov.
\newblock An eigenspace projection clustering method for inexact graph
  matching.
\newblock {\em IEEE transactions on pattern analysis and machine intelligence},
  26(4):515--519, 2004.

\bibitem{calissano2020populations}
Anna Calissano, Aasa Feragen, and Simone Vantini.
\newblock Populations of unlabeled networks: Graph space geometry and geodesic
  principal components.
\newblock 2020.

\bibitem{cour2007balanced}
Timothee Cour, Praveen Srinivasan, and Jianbo Shi.
\newblock Balanced graph matching.
\newblock In {\em Advances in Neural Information Processing Systems}, pages
  313--320, 2007.

\bibitem{dryden2016statistical}
Ian~L Dryden and Kanti~V Mardia.
\newblock {\em Statistical shape analysis: with applications in R}, volume 995.
\newblock John Wiley \& Sons, 2016.

\bibitem{duncan2018statistical}
Adam Duncan, Eric Klassen, Anuj Srivastava, et~al.
\newblock Statistical shape analysis of simplified neuronal trees.
\newblock {\em The Annals of Applied Statistics}, 12(3):1385--1421, 2018.

\bibitem{gold1996graduated}
Steven Gold and Anand Rangarajan.
\newblock A graduated assignment algorithm for graph matching.
\newblock {\em IEEE Transactions on pattern analysis and machine intelligence},
  18(4):377--388, 1996.

\bibitem{guo2019quotient}
Xiaoyang Guo, Anuj Srivastava, and Sudeep Sarkar.
\newblock A quotient space formulation for statistical analysis of graphical
  data.
\newblock {\em arXiv preprint arXiv:1909.12907}, 2019.

\bibitem{hagwood-etal:2013}
Charles Hagwood, Javier Bernal, Michael Halter, John Elliott, and Tegan
  Brennan.
\newblock Testing equality of cell populations based on shape and geodesic
  distances.
\newblock {\em IEEE Transactions on Medical Imaging}, 32(12), 2013.

\bibitem{hang2019topological}
Haibin Hang, Facundo M{\'e}moli, and Washington Mio.
\newblock A topological study of functional data and fr{\'e}chet functions of
  metric measure spaces.
\newblock {\em Journal of Applied and Computational Topology}, 3(4):359--380,
  2019.

\bibitem{hoover2000locating}
AD Hoover, Valentina Kouznetsova, and Michael Goldbaum.
\newblock Locating blood vessels in retinal images by piecewise threshold
  probing of a matched filter response.
\newblock {\em IEEE Transactions on Medical imaging}, 19(3):203--210, 2000.

\bibitem{jain2009structure}
Brijnesh~J Jain and Klaus Obermayer.
\newblock Structure spaces.
\newblock {\em Journal of Machine Learning Research}, 10(Nov):2667--2714, 2009.

\bibitem{jain2011graph}
Brijnesh~J Jain and Klaus Obermayer.
\newblock Graph quantization.
\newblock {\em Computer Vision and Image Understanding}, 115(7):946--961, 2011.

\bibitem{jain2012learning}
Brijnesh~J Jain and Klaus Obermayer.
\newblock Learning in riemannian orbifolds.
\newblock {\em arXiv preprint arXiv:1204.4294}, 2012.

\bibitem{jermyn2012elastic}
Ian~H Jermyn, Sebastian Kurtek, Eric Klassen, and Anuj Srivastava.
\newblock Elastic shape matching of parameterized surfaces using square root
  normal fields.
\newblock In {\em European conference on computer vision}, pages 804--817.
  Springer, 2012.

\bibitem{jermyn2017elastic}
Ian~H Jermyn, Sebastian Kurtek, Hamid Laga, and Anuj Srivastava.
\newblock Elastic shape analysis of three-dimensional objects.
\newblock {\em Synthesis Lectures on Computer Vision}, 12(1):1--185, 2017.

\bibitem{kendall1984shape}
David~G Kendall.
\newblock Shape manifolds, procrustean metrics, and complex projective spaces.
\newblock {\em Bulletin of the London mathematical society}, 16(2):81--121,
  1984.

\bibitem{klassen2004analysis}
Eric Klassen, Anuj Srivastava, M Mio, and Shantanu~H Joshi.
\newblock Analysis of planar shapes using geodesic paths on shape spaces.
\newblock {\em IEEE transactions on pattern analysis and machine intelligence},
  26(3):372--383, 2004.

\bibitem{kong2005diversity}
Jee-Hyun Kong, Daniel~R Fish, Rebecca~L Rockhill, and Richard~H Masland.
\newblock Diversity of ganglion cells in the mouse retina: unsupervised
  morphological classification and its limits.
\newblock {\em Journal of Comparative Neurology}, 489(3):293--310, 2005.

\bibitem{koopmans1957assignment}
Tjalling~C Koopmans and Martin Beckmann.
\newblock Assignment problems and the location of economic activities.
\newblock {\em Econometrica: journal of the Econometric Society}, pages 53--76,
  1957.

\bibitem{kruskal1964multidimensional}
Joseph~B Kruskal.
\newblock Multidimensional scaling by optimizing goodness of fit to a nonmetric
  hypothesis.
\newblock {\em Psychometrika}, 29(1):1--27, 1964.

\bibitem{kurtek2010novel}
Sebastian Kurtek, Eric Klassen, Zhaohua Ding, and Anuj Srivastava.
\newblock A novel riemannian framework for shape analysis of 3d objects.
\newblock In {\em 2010 IEEE Computer Society Conference on Computer Vision and
  Pattern Recognition}, pages 1625--1632. IEEE, 2010.

\bibitem{laga2018survey}
Hamid Laga.
\newblock A survey on nonrigid 3d shape analysis.
\newblock In {\em Academic Press Library in Signal Processing, Volume 6}, pages
  261--304. Elsevier, 2018.

\bibitem{lawler1963quadratic}
Eugene~L Lawler.
\newblock The quadratic assignment problem.
\newblock {\em Management science}, 9(4):586--599, 1963.

\bibitem{leordeanu2005spectral}
Marius Leordeanu and Martial Hebert.
\newblock A spectral technique for correspondence problems using pairwise
  constraints.
\newblock In {\em Tenth IEEE International Conference on Computer Vision
  (ICCV'05) Volume 1}, volume~2, pages 1482--1489. IEEE, 2005.

\bibitem{leordeanu2009integer}
Marius Leordeanu, Martial Hebert, and Rahul Sukthankar.
\newblock An integer projected fixed point method for graph matching and map
  inference.
\newblock In {\em Advances in neural information processing systems}, pages
  1114--1122, 2009.

\bibitem{liu2012extended}
Zhi-Yong Liu, Hong Qiao, and Lei Xu.
\newblock An extended path following algorithm for graph-matching problem.
\newblock {\em IEEE transactions on pattern analysis and machine intelligence},
  34(7):1451--1456, 2012.

\bibitem{revest2009adult}
JM Revest, D Dupret, Muriel Koehl, C Funk-Reiter, N Grosjean, PV Piazza, and DN
  Abrous.
\newblock Adult hippocampal neurogenesis is involved in anxiety-related
  behaviors.
\newblock {\em Molecular psychiatry}, 14(10):959--967, 2009.

\bibitem{shen2014functional}
Dan Shen, Haipeng Shen, Shankar Bhamidi, Yolanda Mu{\~n}oz~Maldonado, Yongdai
  Kim, and J~Stephen Marron.
\newblock Functional data analysis of tree data objects.
\newblock {\em Journal of Computational and Graphical Statistics},
  23(2):418--438, 2014.

\bibitem{singh2014topological}
Nikhil Singh, Heather~D Couture, JS Marron, Charles Perou, and Marc Niethammer.
\newblock Topological descriptors of histology images.
\newblock In {\em International Workshop on Machine Learning in Medical
  Imaging}, pages 231--239. Springer, 2014.

\bibitem{sonnenschein2015image}
Anne Sonnenschein, David VanderZee, William~R Pitchers, Sudarshan Chari, and
  Ian Dworkin.
\newblock An image database of drosophila melanogaster wings for phenomic and
  biometric analysis.
\newblock {\em GigaScience}, 4(1):25, 2015.

\bibitem{srivastava2010shape}
Anuj Srivastava, Eric Klassen, Shantanu~H Joshi, and Ian~H Jermyn.
\newblock Shape analysis of elastic curves in euclidean spaces.
\newblock {\em IEEE Transactions on Pattern Analysis and Machine Intelligence},
  33(7):1415--1428, 2010.

\bibitem{srivastava2016functional}
Anuj Srivastava and Eric~P Klassen.
\newblock {\em Functional and shape data analysis}.
\newblock Springer, 2016.

\bibitem{srivastava2011registration}
Anuj Srivastava, Wei Wu, Sebastian Kurtek, Eric Klassen, and James~Stephen
  Marron.
\newblock Registration of functional data using fisher-rao metric.
\newblock {\em arXiv preprint arXiv:1103.3817}, 2011.

\bibitem{su2019shape}
Zhe Su, Martin Bauer, Stephen~C Preston, Hamid Laga, and Eric Klassen.
\newblock Shape analysis of surfaces using general elastic metrics.
\newblock {\em arXiv preprint arXiv:1910.02045}, 2019.

\bibitem{umeyama1988eigendecomposition}
Shinji Umeyama.
\newblock An eigendecomposition approach to weighted graph matching problems.
\newblock {\em IEEE transactions on pattern analysis and machine intelligence},
  10(5):695--703, 1988.

\bibitem{vogelstein2015fast}
Joshua~T Vogelstein, John~M Conroy, Vince Lyzinski, Louis~J Podrazik, Steven~G
  Kratzer, Eric~T Harley, Donniell~E Fishkind, R~Jacob Vogelstein, and Carey~E
  Priebe.
\newblock Fast approximate quadratic programming for graph matching.
\newblock {\em PLOS one}, 10(4):e0121002, 2015.

\bibitem{wasserman2018topological}
Larry Wasserman.
\newblock Topological data analysis.
\newblock {\em Annual Review of Statistics and Its Application}, 5:501--532,
  2018.

\bibitem{zanfir2018deep}
Andrei Zanfir and Cristian Sminchisescu.
\newblock Deep learning of graph matching.
\newblock In {\em Proceedings of the IEEE Conference on Computer Vision and
  Pattern Recognition}, pages 2684--2693, 2018.

\bibitem{zhang2018phase}
Zhengwu Zhang, Eric Klassen, and Anuj Srivastava.
\newblock Phase-amplitude separation and modeling of spherical trajectories.
\newblock {\em Journal of Computational and Graphical Statistics},
  27(1):85--97, 2018.

\bibitem{zhou2012factorized}
Feng Zhou and Fernando De~la Torre.
\newblock Factorized graph matching.
\newblock In {\em 2012 IEEE Conference on Computer Vision and Pattern
  Recognition}, pages 127--134. IEEE, 2012.

\bibitem{zhou2015factorized}
Feng Zhou and Fernando De~la Torre.
\newblock Factorized graph matching.
\newblock {\em IEEE transactions on pattern analysis and machine intelligence},
  38(9):1774--1789, 2015.

\end{thebibliography}
}

\end{document}